%% file: main.tex
\documentclass[conference]{IEEEtran}
\input{extra_pkgs}

\def\BibTeX{{\rm B\kern-.05em{\sc i\kern-.025em b}\kern-.08em
    T\kern-.1667em\lower.7ex\hbox{E}\kern-.125emX}}

\fancypagestyle{firstpage}{
  \fancyhf{}

  \fancyfoot[C]{\thepage}
}

\title{Echo: Compiler-based GPU Memory Footprint Reduction for LSTM RNN Training} 
\author{}
\author{
    \vspace{-25pt} \\
    Bojian Zheng\(\ssymbol{2}\), Abhishek Tiwari\(\ssymbol{2}\), Nandita Vijaykumar\(\ssymbol{4}\), Gennady Pekhimenko\(\ssymbol{2}\) \\
    \(\ssymbol{2}\)University of Toronto, \(\ssymbol{4}\)Carnegie Mellon University \\
    bojian@cs.toronto.edu, atawari@cs.toronto.edu, nandita@cmu.edu, pekhimenko@cs.toronto.edu \\
    \vspace{-25pt}
}

\begin{document}
\maketitle
\thispagestyle{firstpage}
\pagestyle{plain}


\newcommand{\TODO}      {{\color{red}TODO}}
\newcommand{\CHANGE} [1]{{\color{black}#1}}
\newcommand{\COMMENT}[1]{{\color{red}#1}}
\newcommand{\Baseline}{\emph{Baseline}}
\newcommand{\Mirror}  {\emph{Mirror}}
\newcommand{\Echo}    {\emph{Echo}}
\newcommand{\IWSLTSpeedup}{\(1.28\times\)}
\newcommand{\NMTFootprintReductionRatio}{\(3.13\times\)}
\newcommand{\DSFootprintReductionRatio}{\(1.56\times\)}
\newcommand{\TFFootprintReductionRatio}{\(1.59\times\)}
\newcommand{\TFMaxNLayers}{\(1.83\times\)}
\newcommand{\ResNetFootprintReductionRatio}{\(2.13\times\)}
\newcommand{\ResNetMaxNLayers}{\(4.0\times\)}
\newcommand{\PriorWorkFootprintReductionRatio}{\(2.31\times\)}
\newcommand{\PriorWorkRuntimeOverhead}{\(18\times\)}
\newcommand{\AVGFootprintReductionRatio}{\(1.89\times\)}

\input{sections/0-Abstract}
\input{sections/1-Introduction}
\input{sections/2-Background}
\input{sections/3-Motivation}
\input{sections/4-Key_Ideas}
\input{sections/5-Implementation_Details}
\input{sections/6-Evaluation}
\input{sections/7-Related_Works}
\input{sections/8-Conclusions}


\bibliographystyle{ACM-Reference-Format}
\setlength{\bibsep}{0pt plus 0.3ex}
{\footnotesize
\bibliography{bibliography}
}

\end{document}

%% file: extra_pkgs.tex
\usepackage[T1]{fontenc}
\usepackage{microtype}

\usepackage{nth}
\usepackage{tikz}
\usepackage{xurl}
\usepackage{xcolor}
\usepackage{subcaption}
\usepackage{caption}
\usepackage{mathrsfs}
\usepackage{nicefrac}
\usepackage{listings}
\usepackage{fancyhdr}
\usepackage{anyfontsize}
\usepackage[numbers]{natbib}
\usepackage[export]{adjustbox}
\usepackage[ruled, noend]{algorithm2e}
\usepackage{amsmath, amsthm, amssymb, amsbsy}
\usepackage{booktabs, tabularx, multirow, tablefootnote}

\newcommand*\darkcircled[1]{\tikz[baseline=(char.base)]{
            \node[shape=circle,fill,inner sep=0.5pt] (char) {\textcolor{white}{#1}};}}
\makeatletter
\newcommand\footnoteref[1]{\protected@xdef\@thefnmark{\ref{#1}}\@footnotemark}
\makeatother

\makeatletter
\newcommand{\ssymbol}[1]{^{\@fnsymbol{#1}}}
\makeatother

%% file: sections/0-Abstract.tex
\begin{abstract}
    The Long-Short-Term-Memory Recurrent Neural Networks (LSTM RNNs) are a popular class of machine learning models for analyzing sequential data. Their training on modern GPUs, however, is limited by the GPU memory capacity. Our profiling results of the LSTM RNN-based Neural Machine Translation (NMT) model reveal that feature maps of the attention and RNN layers form the memory bottleneck and runtime is unevenly distributed across different layers when training on GPUs. Based on these two observations, we propose to recompute the feature maps of the attention and RNN layers rather than stashing them persistently in the GPU memory.
    
    While the idea of feature map recomputation has been considered before, existing solutions fail to deliver satisfactory footprint reduction, as they do not address two key challenges.
    For each feature map recomputation to be effective and efficient, its effect on (1) the total memory footprint, and (2) the total execution time has to be carefully estimated.
    To this end, we propose \Echo{}, a new compiler-based optimization scheme that addresses the first challenge with a practical mechanism that estimates the memory benefits of recomputation over the entire computation graph, and the second challenge by non-conservatively estimating the recomputation overhead leveraging layer specifics. \Echo{} reduces the GPU memory footprint \emph{automatically and transparently} without any changes required to the training source code, and is effective for models beyond LSTM RNNs.
    
    
    \CHANGE{We evaluate \Echo{} on numerous state-of-the-art machine learning workloads, including NMT, DeepSpeech2, Transformer, and ResNet, on real systems with modern GPUs and observe footprint reduction ratios of \AVGFootprintReductionRatio{} on average and \NMTFootprintReductionRatio{} maximum. Such reduction can be converted into faster training with a larger batch size, savings in GPU energy consumption (e.g., training with one GPU as fast as with four), and/or an increase in the maximum number of layers under the same GPU memory budget. 
    }

\end{abstract}

%% file: sections/1-Introduction.tex
\section{Introduction}
\label{sec:intro}

LSTM~\cite{lstm} RNNs form an important class of machine learning models for analyzing sequential data, having applications in language modeling~\cite{rnn_regularize, lstm_lm}, machine translation~\cite{nmt, seq2seq, nmt-massive_exploration, gnmt}, and speech recognition~\cite{deepspeech, end2end-speech}. In these tasks, LSTM RNNs are trained over large amounts of data samples (e.g., sentences or audio files) to capture their inherent temporal dependencies.
Despite their importance, LSTM RNN training tends to be less efficient on modern GPUs compared to other types of deep neural networks (DNNs) such as Convolutional Neural Networks (CNNs) \cite{sru, quasi}. 
One of the main reasons for such inefficiency is the high GPU memory consumption of LSTM RNNs that limits the maximum training batch size, which, in turn, limits the GPU compute utilization because of the small amount of available data parallelism \cite{tbd}.

There have been numerous works \cite{eie, deep_compression, vdnn, gist} that propose techniques for memory footprint reduction in DNNs, but these works, unfortunately, have limited applicability for LSTM RNN training. Specifically, prior works that propose efficient compression techniques for inference (e.g., \cite{eie, deep_compression}) focus on weights rather than the feature maps (which consume the majority of the overall memory in DNN training \cite{vdnn, gist}). Prior works such as vDNN~\cite{vdnn} and Gist~\cite{gist} that attempt to reduce footprint in CNN training cannot be directly applied to LSTM RNNs as they would either lead to (i) high runtime overhead for many small vector layers used in LSTM RNNs, or (ii) limited applicability, as LSTM RNNs use \(\tanh\)/\(\operatorname{sigmoid}\), rather than \(\operatorname{ReLU}\) activations, resulting in almost no opportunities for the data encodings proposed in \emph{Gist}~\cite{gist}.

To better understand the reasons that lead to the high GPU memory consumption in LSTM RNNs, we perform a detailed breakdown analysis of the GPU memory consumption (and complimentary runtime analysis) during the training of the state-of-the-art LSTM RNN-based NMT model \cite{nmt, gnmt}. We observe that (i) the feature maps of the attention and RNN layers consume the majority of the GPU memory (\emph{feature maps} are intermediate data entries saved in the forward pass to compute the gradients during the backward pass), and 
(ii) the runtime is unevenly distributed across different layers (fully-connected layers dominate the runtime while other layers are relatively lightweight).
From these two observations, we adopt the idea of \emph{selective recomputation}, where we can leverage the low computational cost of non-fully-connected layers to \emph{recompute} their feature maps during the backward pass, rather than stashing them in the GPU memory \cite{sublinear, memory_effi-bptt, hessian-free}.

Although the idea of feature map recomputation has been explored before in prior works \cite{sublinear, memory_effi-bptt, hessian-free}, they fail to deliver satisfactory footprint reduction in the case of LSTM RNN training and other state-of-the-art training workloads (as we will show 
in Sections~\ref{subsec:motivation:backward_mirroring}~and~\ref{sec:evaluation}). These previous proposals are ineffective in the LSTM RNN context as a result of not addressing two important challenges:


(1) \emph{Accurately estimating footprint reduction.} 
While recomputation obviates the need to store feature maps of some layers (or group of layers),
it needs to \emph{additionally} stash some data entries that are needed to recompute these (group of) layers as new feature maps.
Hence, a practical recomputation strategy should involve the comparison between the feature maps that are released and the ones that are newly allocated, but such a comparison is far from being trivial and is overlooked by prior works \cite{sublinear}. First, when making a decision on whether an operator should be recomputed, we should focus not only on the storage allocations that are \emph{local} to this operator, but also on the \emph{global effects} of these allocations, and take into account any potential \emph{reuse} across different operators within the same computation graph. Second, we need a practical mechanism to estimate the recomputation benefits over the entire graph. Na\"ive implementation has quadratic runtime complexity, which can be impractical in the LSTM RNN context given that the number of the operator nodes in the graph is usually huge (e.g., more than \(10000\) in large NMT models \cite{sockeye}).

(2) \emph{Non-conservatively estimating runtime overhead.} 
Recomputation always comes with runtime overhead (as feature maps have to be recomputed), and hence potential targets for recomputation have to be carefully selected. Na\"ive implementations simply exclude any compute-heavy layers (e.g., convolutions, fully-connected layers) to keep the recomputation overhead low \cite{sublinear}. Such an approach \CHANGE{is} too conservative and leads to limited applicability in the LSTM RNN context. 
We, however, notice that if \emph{layer specifics} are taken into account, certain layers that are otherwise filtered out can be amenable to recomputation with low overhead. 
For example, the fully-connected layers do not need recomputation as their gradients' computation does not need their outputs. Other examples of layer specifics include binarization~\cite{gist} for \(\operatorname{ReLU}\) activations and \(\operatorname{dropout}\) layers~\cite{dropout}. Those layers require special handling so that we would not miss opportunities for footprint reduction and still avoid significant runtime overhead. 

To effectively address these challenges and enable practical recomputation in LSTM RNN training, we propose \emph{Echo}, a new compiler-based optimization scheme.
\emph{Echo} employs two key ideas to achieve this goal. To address the first challenge, \emph{Echo} makes footprint reduction estimation practical by partitioning the whole computation graph into smaller subgraphs to restrict the scope (and hence reduce the complexity) of the footprint reduction estimation. Compute-heavy layers form the natural boundaries for partition, since they are not recomputed and therefore out of the estimation scope.
\Echo{} then analyzes each small subgraph \emph{independently} and makes accurate footprint reduction estimation for recomputation.
To address the second challenge, \Echo{} infers the data dependencies of the gradient operators. \emph{Only if} the gradient computation requires the forward operators' outputs will the forward operators' runtime be added as a part of the recomputation overhead estimate.

Our major contributions can be summarized as follows:

(1) We present a detailed breakdown and analysis of how the GPU memory is consumed and where the runtime is spent in NMT training. Our profiling reveals that the feature maps of the attention and RNN layers form the memory bottleneck and the runtime is unevenly distributed across different layers, which motivates us to adopt the idea of selective recomputation to reduce the GPU memory footprint.

(2) \CHANGE{We find and address the key challenges in making selective recomputation practical by carefully estimating the footprint reduction and runtime overhead, therefore significantly outperforming prior works in both aspects.}

(3) We implement our ideas in a new graph optimization pass, \emph{Echo}, that is a part of the graph compiler of the machine learning framework (MXNet~\cite{mxnet} NNVM~\cite{nnvm}). \emph{Echo} reduces the GPU memory footprint \emph{transparently and automatically} for numerous state-of-the-art machine learning models without any changes to the training source code (even beyond LSTM RNNs). As Section~\ref{sec:evaluation} will show, we \CHANGE{additionally implement} hand-tuned CUDA kernels that perform recomputation more efficiently but those kernels take us several weeks to develop for every specific model, \CHANGE{whereas \Echo{} is effective for all the models we have examined in a fully automatic way}.

(4) We evaluate \emph{Echo} in a state-of-the-art machine learning framework (MXNet~\cite{mxnet}) on four state-of-the-art machine learning models \cite{mlperf, tbd} used for machine translation (NMT~\cite{nmt, gnmt}, Transformer~\cite{transformer}), speech recognition (DeepSpeech2~\cite{deepspeech2}), and image classification (ResNet~\cite{resnet}), and observe GPU memory footprint reduction ratio of \NMTFootprintReductionRatio{}, \DSFootprintReductionRatio{}, \TFFootprintReductionRatio{}, and \ResNetFootprintReductionRatio{} correspondingly. On the NMT model, we demonstrate that this reduction can be converted into training to the same quality with a larger batch size \IWSLTSpeedup{} faster or training with one GPU as fast as with four, \CHANGE{and on Transformer and ResNet models, we further show that our approach can help increase the maximum number of layers by \TFMaxNLayers{} and \ResNetMaxNLayers{} respectively while using the same GPU memory budget.}

%% file: sections/2-Background.tex
\section{Background}
\label{sec:background}

This section gives a short overview of LSTM RNNs and their applications for machine translation tasks. For simplicity, we hide the algorithmic details that are not relevant to this work and focus on the tensor shape transformations across different layers that matter most in memory allocations. \textbf{Bold text} in this section will be used as examples in Section~\ref{sec:key_ideas}.

\begin{figure}[!t]
    \centering
    \includegraphics[width=\linewidth]{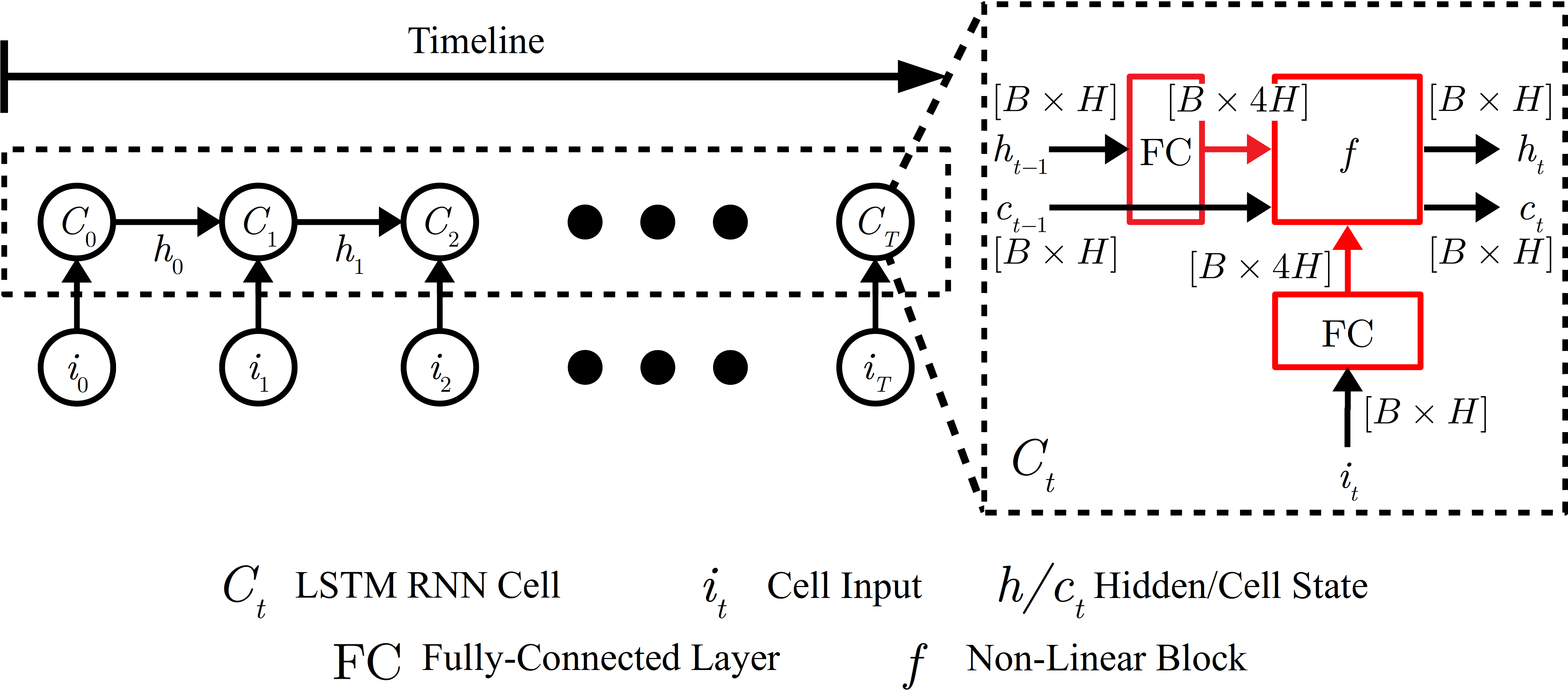}
    \captionsetup{width=0.95\linewidth}
    \caption{Left: A single-layer LSTM RNN that scans through an input sequence. Right: Zoom-in view of one LSTM cell. Both diagrams are simplified for clarity.}
    \label{fig:lstm_rnn}
    \vspace{-2pt}
\end{figure}

Figure~\ref{fig:lstm_rnn} shows a simplified view of a single-layer LSTM RNN that reads through an input sequence \(i_{1\sim T}\), where the annotations in square brackets denote the tensor shapes. The layer has \(T\) LSTM cells \(C_{1\sim T}\), where \(T\) is the input sequence length (e.g., in the context of machine translation, \(T\) denotes the number of words per sentence). Each cell \(C_t\) receives three inputs from two directions: (1) from the input \(i_t\) of the current time step, and (2) from the hidden and cell state of the previous cell. All inputs are of dimension \([B\times H]\), where \(B\) is the batch size and \(H\) is the hidden dimension. Both \(i_t\) and \(h_t\) need to go through a fully-connected layer, defined by Equation~(\ref{eqn:fc}):

\vspace{-\baselineskip}
\begin{equation}\label{eqn:fc}
    Y=XW^T+b, W: [4H\times H], b: [4H]
\end{equation}
\vspace{-1.5\baselineskip}

where \(X, Y, W, b\) are input, output, weight, and bias respectively. The weight and bias are shared across the timeline.

After becoming \(4\times\) larger in terms of hidden dimension by Equation~\ref{eqn:fc}, \(i_t\) and \(h_t\), together with \(c_t\), enter the non-linear block \(f\) that consists of slicing and element-wise operations. The output of the LSTM cell is the hidden and cell state of the current time step, both of which are of dimension \([B\times H]\). 
\textbf{The sum of the non-linear block's input sizes \([9\times B\times H]\) is greater than its output sizes \([B\times H]\), which will become an example in Section~\ref{subsec:key_ideas:footprint} (Figure~\ref{fig:add_and_tanh}}).

\begin{figure}[!t]
    \centering
    \includegraphics[width=\linewidth]{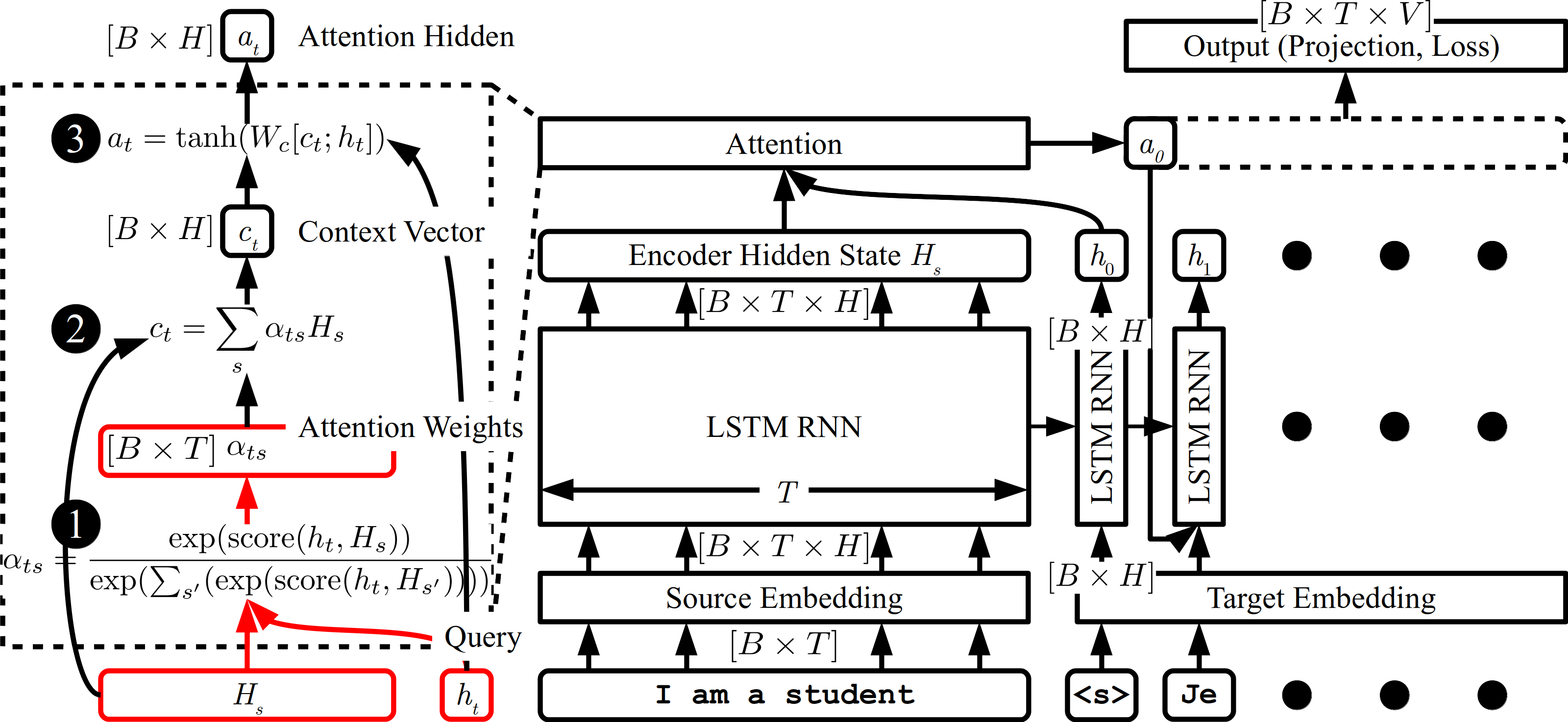}
    
    \captionsetup{width=\linewidth, justification=centering}
    \caption{NMT (LSTM RNN Encoder (Middle)-Decoder (Right) with Attention (Left)
    )}
    \label{fig:nmt}
    \vspace{-2pt}
\end{figure}

The NMT model~\cite{nmt, gnmt} is the state-of-the-art LSTM RNN-based model used for machine translation. It has three major blocks, namely the encoder, decoder, and attention (Figure~\ref{fig:nmt}). 
In the encoder, source sentences of the training dataset are batched into a tensor of shape \([B\times T]\). In the embedding layer, each word in a sentence is encoded into a hidden state of dimension \(H\). The result is sent to the LSTM RNN as an input (see Figure~\ref{fig:lstm_rnn}). The hidden states of the LSTM cell at all time steps (each of which is of shape \([B\times H]\)) are concatenated together into the source hidden state (\([B\times T\times H]\)).
The encoder passes its internal hidden states to the decoder, which decodes the target sentences one word at a time. It decodes one target word into a hidden state \(h_t\) (\([B\times H]\)) also known as a query.

The query and encoder hidden state are given to the attention layer, where they go through the following procedures to generate the attention hidden state \(a_t\) (see Figure~\ref{fig:nmt}): 

\darkcircled{1} A scoring function compares the query with the encoder hidden state, generating the attention scores (\([B\times H]\)), which is used to determine the attention weights \(\alpha_{ts}\) (\([B\times H]\)). \textbf{The encoder hidden state \(H_s\) is reused across all the time steps, which will become an example in Section~\ref{subsec:key_ideas:footprint} (Figure~\ref{fig:mlp_mirror_local}}).

\darkcircled{2} A context vector \(c_t\) (\([B\times H]\)) is computed as the attention weights-weighted average of the encoder hidden state.

\darkcircled{3} The query and context vector are concatenated together to generate the attention hidden state \(a_t\) (\([B\times H]\)), 
which is sent to the next decoder time step for the next word in sequence.

This process continues until the maximum sequence length is reached. The decoded sequence is then sent to the output layer to generate the training loss. In training, the loss is first computed during the forward pass, and then propagated back through the network to compute the gradients that are used to update the model weights. By the nature of the backpropagation algorithm \cite{backprop}, some data entries have to be stashed in memory during the forward pass to compute the gradients \cite{gist}.

%% file: sections/3-Motivation.tex
\section{Motivation}
\label{sec:motivation}

\subsection{Why does GPU memory footprint matter?}
\label{subsec:motivation:why}

There are two major benefits of GPU memory footprint reduction. First comes from boosting the training performance by using larger training batch size,\footnote{Although one might argue that large training batch size might hurt convergence, in Section~\ref{subsec:evaluation:nmt} we show actual training curves that increase convergence speed to the same quality when training with larger batch size.} and second from allowing to train wider and deeper models with the same GPU resources.

\begin{figure}[ht]
    \centering
    \begin{subfigure}[b]{0.49\linewidth}
        \includegraphics[width=\linewidth]{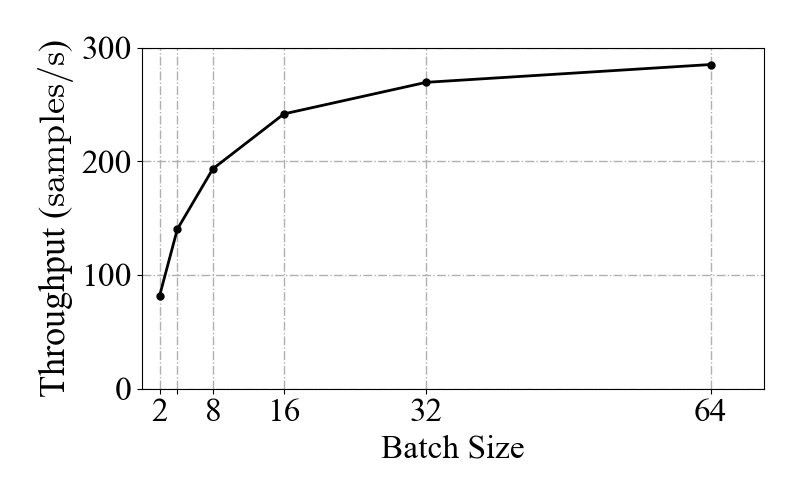}
        \vspace{-20pt}
        \caption{ResNet-50}
        \label{subfig:resnet_50}
    \end{subfigure}
    \begin{subfigure}[b]{0.49\linewidth}
        \includegraphics[width=\linewidth]{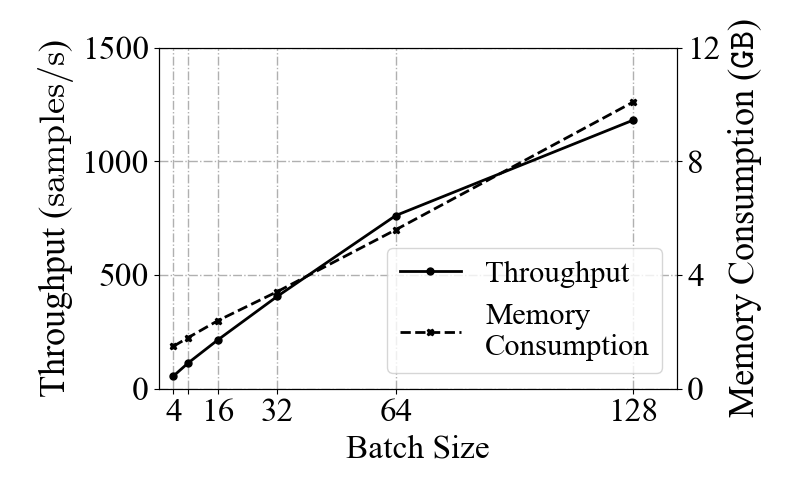}
        \vspace{-20pt}
        \caption{NMT}
        \label{subfig:sockeye}
    \end{subfigure}
    \captionsetup{width=0.98\linewidth}
    \caption{(a) Training Throughput of ResNet-50 versus Batch Size (b) Training Throughput and GPU Memory Usage of NMT versus Batch Size (using one RTX 2080 Ti GPU
    )}
    \label{fig:resnet_50-vs-sockeye}
\end{figure}

\textbf{Increase Training Throughput with Larger Batch Size.} 
We compare the training throughput between ResNet-50~\cite{resnet} (CNN-based model used for image classification) and NMT~\cite{nmt, gnmt} (LSTM RNN-based model) with respect to their training batch size. Figure~\ref{subfig:resnet_50} shows the correlation between training throughput (measured as \(\mathrm{samples}/\mathrm{second}\)) and batch size for ResNet-50 (detailed methodology in Section~\ref{subsec:evaluation:methodology}). We notice that the training throughput saturates as the batch size increases. Previous work by \citet{tbd} reveals that the GPU compute units have been almost fully utilized (starting from batch size of \(32\)) and therefore further increasing the batch size yields little benefit on the training throughput. However, the story is different in LSTM RNNs. Figure~\ref{subfig:sockeye} shows a similar graph for NMT. We observe that the training throughput increases linearly with the batch size, but such increase stops when the model hits the GPU memory capacity wall on a modern \(11\ \mathtt{GB}\) RTX 2080 Ti GPU~\cite{2080-ti} at the batch size of \(128\), and cannot increase any further. From the comparison, we draw the conclusion that, \textbf{in LSTM RNN-based model, performance is limited by the GPU memory capacity}, and hence this justifies why footprint reduction techniques can further increase the training throughput for such models (as we will show in Section~\ref{subsec:evaluation:nmt}). 

\textbf{Run Wider and Deeper Models.} Although models such as ResNet might not always be able to benefit from the footprint reduction to achieve performance gains, they can still benefit indirectly by becoming \textbf{wider and deeper} while using the same GPU memory budget. For example, data encoding approach such as \emph{Gist} can increase the maximum number of layers in ResNet from \(851\) to \(1202\) with a batch size of \(16\)~\cite{gist}. In fact, as deep learning models grow larger, recent years have seen models that cannot fit into a single GPU even with a batch size of \(1\)~\cite{google-bert, 3d_conv}. All these models can become more practical if efficient footprint reduction techniques are applied.

\subsection{Memory Consumption Breakdown}
\label{subsec:motivation:memory}

To understand the reason behind NMT's large GPU memory footprint, we develop a memory profiler to show the detailed breakdown. Our analysis is based on two orthogonal ways of categorizing memory: (1) by layer types (e.g., RNN), and (2) by data structures. Major data structures include:

(1) \textbf{Feature Maps}: Each layer needs memory for its own input and output variables. If any of those variables are needed in the backward pass to compute the gradients, it is stashed persistently in memory as feature maps, while those that are not can be released back to the storage pool. For example, consider the \(\tanh\) activation function \(\smash{Y=\tanh(X)}\). Since we have \(\smash{Y'=1-\tanh^2(X)}\), the value of \(\smash{\tanh(X)}\) needs to be stored for the gradient computation in the backward pass.

(2) \textbf{Weights}: Layers such as fully-connected layers (Equation~\ref{eqn:fc}) have parameters \(W, B\) that are optimized as training progresses. In the following text, we use the term \emph{Weights} as a generic term that includes \(W, B\), plus their respective gradients and optimizer states which are used to do weight updates.

(3) \textbf{Workspace} is the scratchpad of a layer to compute the results. When a layer completes its own forward or backward pass, its workspace, if previously requested, can be freed.


\begin{figure}[ht]
    \centering
    \includegraphics[height=120pt]{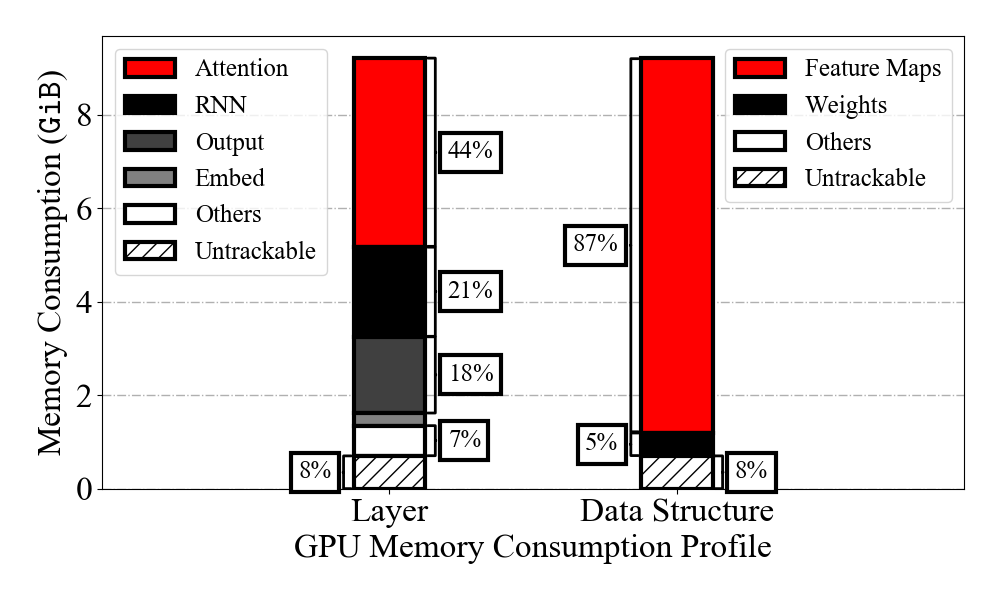}
    \vspace{-5pt}
    \captionsetup{justification=centering}
    \caption{Memory Consumption Breakdown by \\ Layer Types (Left) and Data Structures
    (Right)}
    \label{fig:sockeye-memory_profile}
\end{figure}

Figure~\ref{fig:sockeye-memory_profile} shows the NMT memory consumption breakdown, with the left bar classifying memory consumption by layer types and the right by data structures. The two striped bars at the bottom show the discrepancy between the total amount of memory consumption reported by the memory profiler versus the actual memory usage given by the \emph{nvidia-smi} tool~\cite{nvidia-smi}. Such discrepancy can be caused by memory fragmentation or allocations by the CUDA libraries \cite{mxnet-memory_behavior}. We conclude from the figure that \textbf{the feature maps of the attention layers, followed by RNN and output, are the major memory consumers}, as they are responsible for a total of \(87\%\) (\(8.0\ \mathtt{GB}\)) of the GPU memory used by NMT~\cite{nmt, gnmt}.

\subsection{Runtime Breakdown}
\label{subsec:motivation:runtime}

\begin{figure}[ht]
    \begin{minipage}{0.49\linewidth}
        \includegraphics[height=80pt]{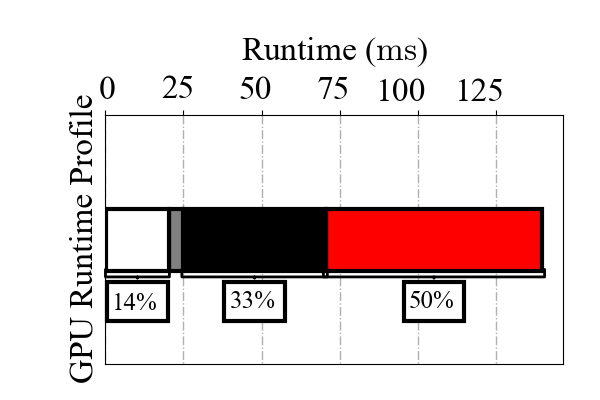}
    \end{minipage}
    \begin{minipage}{0.49\linewidth}
        \begin{minipage}{0.72\linewidth}
            \caption{Runtime Breakdown by Layer Types}
            \label{fig:sockeye-runtime_profile}
        \end{minipage}
    
        \includegraphics[width=0.5\linewidth]{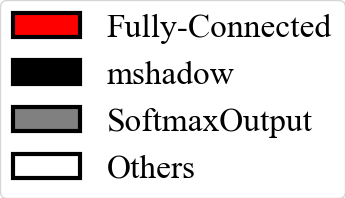}
    \end{minipage}
    \vspace{-5pt}
\end{figure}

To motivate the use of selective recomputation approach, we do a runtime profile analysis that shows the runtime distribution across different layers. 
Figure~\ref{fig:sockeye-runtime_profile} shows the NMT runtime breakdown on one training iteration. The profile is obtained from the \emph{NVProf} tool~\cite{nvprof}. We observe that \textbf{the runtime is  unevenly distributed across different layers}, with \(50\%\) going into the fully-connected layers and the other \(50\%\) into many small compute kernels. The longest kernel of the latter runs for only \(5\text{\ ms}\) (the \emph{mshadow} bar represents the tensor library backend of MXNet and consists of multiple CUDA kernels).


\subsection{Selective Recomputation}
\label{subsec:motivation:backward_mirroring}

Given that the GPU memory consumption limits training performance and that the execution time is distributed unevenly across different layers,
selective recomputation~\cite{sublinear, memory_effi-bptt, hessian-free} can be a viable option to trade runtime with memory capacity.

\begin{figure}[ht]
    \centering
    \vspace{-2pt}
    
    \begin{subfigure}[b]{0.9\linewidth}
        \includegraphics[width=0.99\linewidth, right]{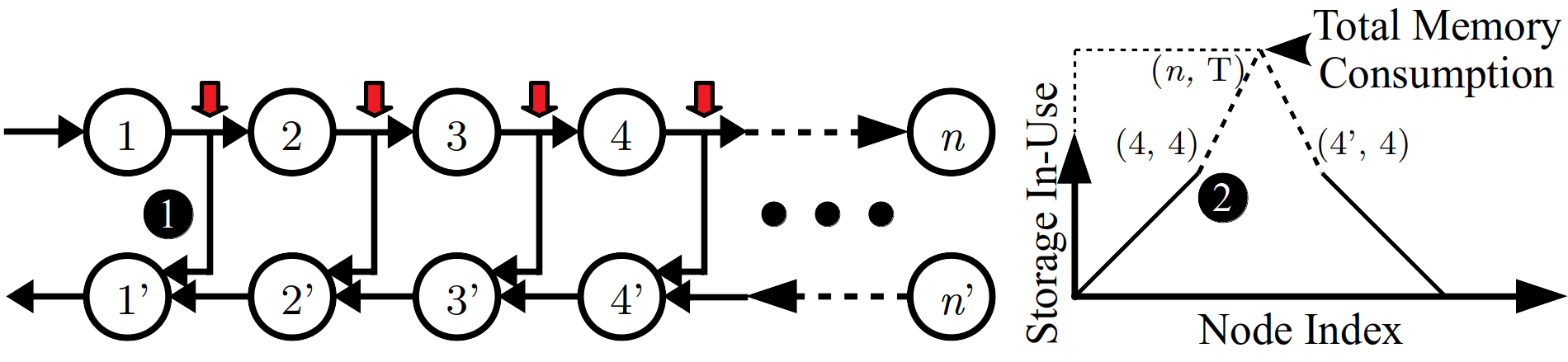}
        \vspace{-20pt}
        \caption{Baseline}
    \end{subfigure}

    \vspace{-8pt}

    \begin{subfigure}[b]{0.91\linewidth}
        \includegraphics[width=\linewidth]{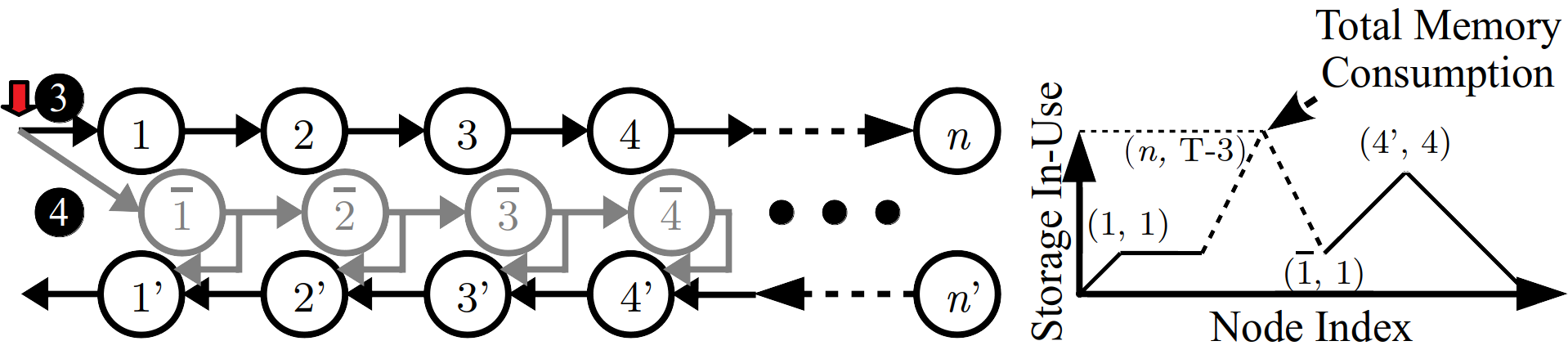}
        \vspace{-20pt}
        \caption{Recomputation}
        \label{subfig:4_tanh_nodes:mirror}
    \end{subfigure}
    \captionsetup{justification=centering}
    \caption{A Computation Graph with Recomputation Applied
        (Red Arrows denote Persistent Feature Maps Storage)}
    \label{fig:4_tanh_nodes}
\end{figure}

To illustrate the key idea behind selective recomputation, consider the computation graph in Figure~\ref{fig:4_tanh_nodes}, where we have a sequence of \(n\) operator nodes. Assume that the gradient node \(i'\) on the backward pass has data dependency on the output edge of its corresponding forward node \(i\). 
The dependency is shown as an edge pointing from the forward to the backward pass (\darkcircled{1}), which is marked as the feature map that has to be stashed. Therefore, the memory allocated for those edges cannot be released back to the storage pool, resulting in four persistent storage units by the time the forward pass completes (\darkcircled{2}). If recomputation is used, those four dependency edges can be replaced with only one edge on the input to Node \#\(1\) (\darkcircled{3}). This releases storage pressure as the inputs to Node \#\(2\sim 4\) can now be taken by their respective outputs (and hence do not need to be stashed anymore), but it comes with the cost of having to redo the forward computation when the backward pass starts (\darkcircled{4}). 
This certainly comes with runtime overhead, but such overhead can be controlled if the recomputed nodes (shown in gray in Figure~\ref{subfig:4_tanh_nodes:mirror}) are restricted to those that are computationally cheap. Hence, selective recomputation has the potential to reduce the memory footprint at small runtime cost.

In practice, however, we observe that the current state-of-the-art recomputation approach has limited benefits on the NMT training. Table~\ref{tab:iwslt15-vi_en-mirror} compares the training performance and GPU memory footprint between training with and without selective recomputation, where the recomputation implementation is based on the prior work~\cite{sublinear}. We observe that recomputation causes the performance to drop by \(17\%\), and it can only give a footprint reduction of \(27\%\) in return, which does not provide enough memory space to increase the performance by, for example, increasing the batch size.

\vspace{2pt}
\begin{table}[ht]
    \centering
    \footnotesize
    \begin{tabular}{lcc}
        \toprule
         & \textbf{Baseline} & \textbf{\citet{sublinear}} \\
        \midrule
        \textbf{Avg. Throughput (\(\mathrm{samples}/\mathrm{s}\))}
            & \(1192\) & \(983\) \\
        \textbf{GPU Memory Footprint (\(\mathtt{GB}\))} 
            & \(10.0\) & \(7.4\) \\
        \bottomrule
    \end{tabular}
    \captionsetup{justification=centering, width=1.1\linewidth}
    \caption{NMT \cite{sockeye} Training with \& without Recomputation}
    \label{tab:iwslt15-vi_en-mirror}
\end{table}



%% file: sections/4-Key_Ideas.tex
\section{Key Ideas}
\label{sec:key_ideas}

The major reason for the ineffectiveness of the state-of-the-art implementations of the recomputation approach is because they fail to adequately address two important challenges:

\subsection{Footprint Reduction Estimation}
\label{subsec:key_ideas:footprint}

\emph{Challenge \#1. Accurately estimating footprint reduction.}
The effect of each operator recomputation on the total memory footprint needs to be carefully estimated. We observe from the red arrows in Figure~\ref{subfig:4_tanh_nodes:mirror} that although recomputation removes the feature maps on the outputs of Node \(\#1\sim 4\), it also brings in a new data entry which is the input to Node \(\#1\) that now has to be stashed. In real models, it is possible for the latter to be larger than the former combined.

\begin{figure}[ht]
    \centering
    \vspace{-5pt}
    \begin{minipage}{0.65\linewidth}
        \begin{subfigure}[b]{0.48\linewidth}
            \includegraphics[width=\linewidth]{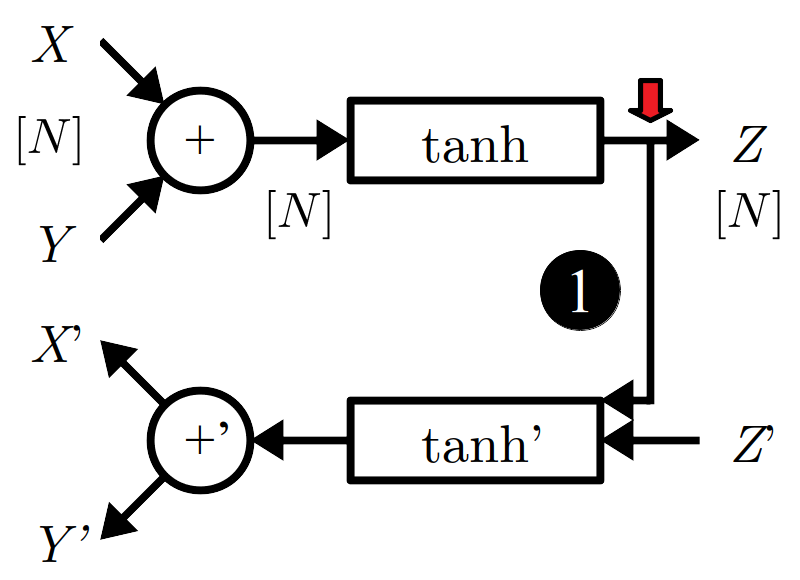}
            \vspace{-15pt}
            \caption{Baseline}
            \label{subfig:add_and_tanh}
        \end{subfigure}
        \begin{subfigure}[b]{0.50\linewidth}
            \includegraphics[width=\linewidth]{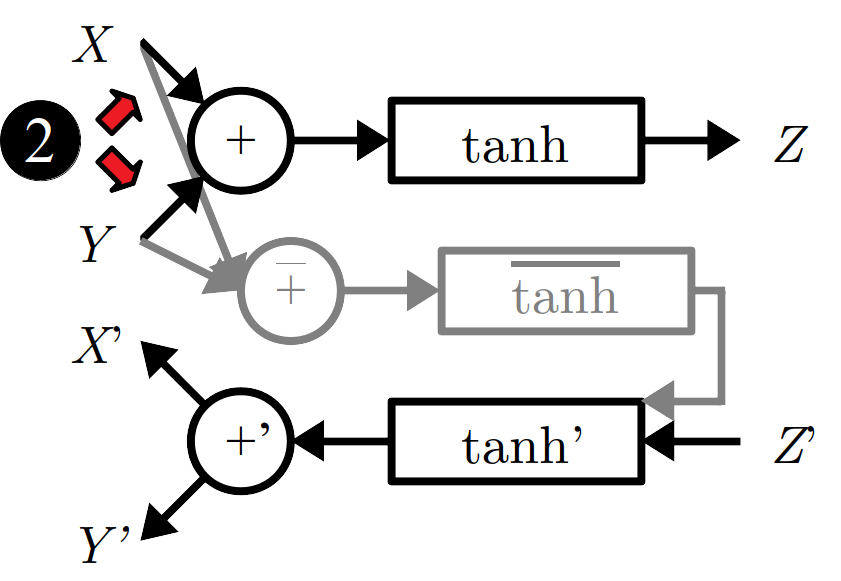}
            \vspace{-15pt}
            \caption{Recomputation}
            \label{subfig:add_and_tanh:mirror}
        \end{subfigure}
    \end{minipage}
    \begin{minipage}{0.33\linewidth}
        \captionsetup{justification=centering}
        \caption{\(Z=\tanh(X+Y)\)}
        \label{fig:add_and_tanh}
    \end{minipage}
\end{figure}

Consider a more concrete example in Figure~\ref{subfig:add_and_tanh}, where \(X\) and \(Y\), both being 1D arrays of shape \([N]\), are elementwise-added first before passing through a \(\tanh\) activation function. The output \(Z\) is also of shape \([N]\). This example is a simplified version of the LSTM cell described in Section~\ref{sec:background} (Figure~\ref{fig:lstm_rnn}).

In the baseline, the framework only stashes \(Z\) as the feature map (\darkcircled{1}), because it is the only value that is needed to compute the gradient in the backward pass.
With the \(+\) and \(\tanh\) listed as recomputed in Figure~\ref{subfig:add_and_tanh:mirror}, the backward dependency is brought forward from \(Z\) to \(X\) and \(Y\) (\darkcircled{2}), similar to Figure~\ref{subfig:4_tanh_nodes:mirror}. However, the only memory saving is from the  feature map \(Z\), while the new ones (\(X\) and \(Y\)) need new allocations, doubling the amount of persistent storage required from \(N\) to \(2N\).

We therefore conclude that a practical recomputation strategy should involve the comparison between the feature maps that are released and those that are newly allocated. However, such a comparison is far from being trivial and is overlooked by prior works \cite{sublinear}. Although a simple comparison between the inputs size and outputs suffices to do the job in the example above, it only considers the storage allocations that are \emph{local} to each operator and ignores the \emph{global impact} of those allocations as it fails to consider the \emph{storage reuse} across different operators within the same computation graph. 

\begin{figure}[ht]
    \centering
    \begin{minipage}{0.33\linewidth}
        \includegraphics[width=\linewidth]{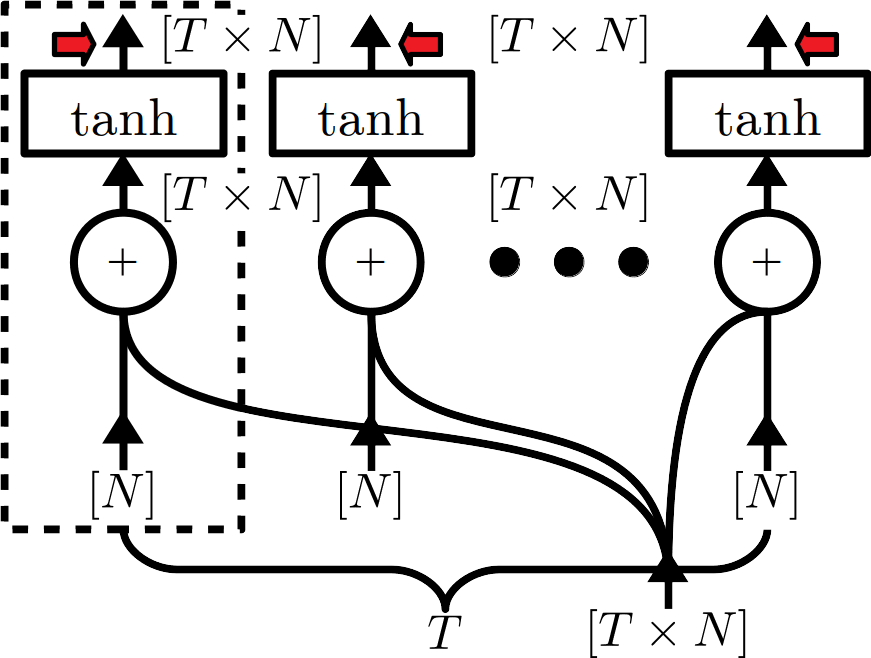}
    \end{minipage}
    \begin{minipage}{0.65\linewidth}
        \captionsetup{justification=centering}
        \caption{\(T\) Tensors of Shape \([N]\) Broadcast-Added with \\
            \([T\times N]\) and \(\tanh\)-Activated}
        \label{fig:mlp_mirror_local}
    \end{minipage}
\end{figure}

Consider another example in Figure~\ref{fig:mlp_mirror_local}, where we have \(T\) tensors of shape \(N\) broadcast-added with the same tensor of shape \([T\times N]\) and then \(\tanh\)-activated. The example is a simplified version of the attention scoring function introduced in Section~\ref{sec:background} Figure~\ref{fig:nmt}. If we restrain the analysis scope within the dashed box and na\"ively compare the inputs size versus outputs, we will arrive at the same conclusion as Figure~\ref{fig:add_and_tanh} that recomputation is not needed. However, with recomputation the total feature map size can be reduced from \(T^2\times N\) to \(T\times 2N\) and the key reason is because the storage of \([T\times N]\) is shared by multiple operators. Therefore, 
we conclude that a global graph analysis is needed to take \emph{reuse} effects into account when using recomputation.

A simple but impractical solution could be an entire computation graph traversal to verify the memory benefits of recomputing for every operator in the graph. This solution has\penalty -10000 quadratic runtime complexity, which is challenging in the LSTM RNN context because the number of operators in the graph is usually huge (e.g., more than \(10\mathrm{K}\) in large NMT models \cite{sockeye}).
To address this challenge, we propose to partition the whole computation graph into small subgraphs to restrict the scope (and hence reduce the complexity) of the footprint reduction analysis. Compute-heavy layers (e.g., fully-connected layers) form the natural boundaries. This is because the goal of the footprint reduction analysis is to estimate the effect of \emph{recomputation} on the total memory footprint. Since compute-heavy layers are never recomputed to avoid significant runtime overhead (as we mention in Section~\ref{subsec:motivation:backward_mirroring}), they are excluded from the scope of analysis and hence can serve as a good place to partition. After the partitioning, each subgraph captures any \emph{reuse} across tensor edges in the computation graph. We then analyze each small subgraph \emph{independently} and accurately estimate potential footprint reduction from recomputation.

In Section~\ref{subsec:impl_details:footprint}, we show examples how footprint reduction estimation can help to (i) avoid pathological cases such as Figure~\ref{fig:add_and_tanh} where recomputation increases the memory footprint, and (ii) reduce the recomputation runtime overhead.


\subsection{Runtime Overhead Estimation}
\label{subsec:key_ideas:runtime}

\emph{Challenge \#2. Non-conservatively estimating runtime overhead.}
The effect of every operator recomputation on the total execution time needs to be carefully estimated taking layer specifics into consideration. From Figure~\ref{subfig:4_tanh_nodes:mirror}, we observe that recomputation always comes with runtime overhead, because feature maps have to be recomputed in the backward pass. Therefore, in practical implementations recomputation is always done selectively. One na\"ive way to do this is to simply exclude all the feature maps of the compute-heavy operators (e.g., convolutions, fully-connected) from being recomputed to keep the runtime overhead low \cite{sublinear}. This approach, however, is too conservative and leads to limited applicability in the LSTM RNN context. A common example is the fully-connected layer that we have listed in Section~\ref{sec:background} Equation~\ref{eqn:fc}:

\vspace{-\baselineskip}
\begin{equation}\label{eqn:fc_bwd}
    Y=XW^T\Rightarrow \frac{dE}{dX}=\frac{dE}{dY}W, \frac{dE}{dW}=\frac{dE}{dY}^T X
\end{equation}
\vspace{-1.2\baselineskip}

where \(E\) is the training loss that has to be optimized.

We observe that the gradients of the fully-connected layer only have data dependency on \(X\) and \(W\) (but not \(Y\)), both of which are the \emph{inputs} to the fully-connected layer. This implies that to recompute the feature maps of the fully-connected layer one does not have to recompute the layer itself (as the output \(Y\) is not needed).
Such a property is layer specific. There are also\penalty -10000 other types of layer specific properties that can enable efficient footprint reduction. For example, the feature maps of \(\operatorname{ReLU}\) activations \cite{gist} and \(\operatorname{dropout}\) layers \cite{dropout} can be stored in \(1\)-bit binary format. Those layers require special handling so that we do not miss opportunities for footprint reduction and still avoid significant runtime overhead.


To address this challenge, we design a non-conservative runtime overhead estimator that infers the layer specifics before estimating the recomputation overhead. 
In the context of fully-connected layers, the idea is that the gradient computation of these layers only need the inputs rather than the outputs.
The estimator's goal is therefore to distinguish between those two cases, and \emph{only if} the feature maps of an operator are part of its outputs will the operator's runtime be added as a part of the overhead estimate. The key challenge for the estimator design lies in its generality. Although hard coding the layer specifics is possible, it is not a scalable solution given the number of layers supported in a machine learning framework \cite{mxnet}. Hence, proper dataflow analysis is needed, as we show in Section~\ref{subsec:impl_details:dead_node}.

\subsection{Automatic and Transparent Compiler Pass Design}
\label{subsec:key_ideas:compiler}

Machine learning programmers always have the option to do recomputation manually by writing hand-tuned GPU kernels. As Section~\ref{sec:evaluation} will show, manual recomputation has the advantage of greatly reducing the recomputation overhead (and even improve runtime due to fewer memory accesses).

Despite its merits, manual recomputation has the following major shortcomings: (i) it is hard to pinpoint the correct places where recomputation should be done, especially in an LSTM RNN model with more than \(10000\) operators \cite{sockeye}, and (ii) it requires nontrivial knowledge of GPU programming, machine learning algorithms, and framework system integration. 
This places significant burden on the programmer.
We first added the recomputation optimization manually to the NMT model and spent more than two weeks seeking recomputation opportunities and testing the manual implementations in MXNet~\cite{mxnet}. This non-trivial effort hence motivates us to push for an \emph{automatic and transparent} method to perform this optimization (and we will show the performance comparison of manual versus automatic version in Section~\ref{subsec:evaluation:nmt}). An automatic and transparent recomputation scheme should reduce the GPU memory footprint without any changes needed to the training source code, while guaranteeing that (i) the recomputation impact on training performance is minimal and (ii) the GPU memory footprint is never worse than that in the baseline.

To this end, we present \Echo{}, a compiler-based optimization scheme that can reduce the GPU memory footprint automatically and transparently. 
\Echo{} addresses the two key challenges for accurate footprint reduction and non-conservative runtime overhead estimation in a computation graph compiler middle-end, without leveraging domain-specific knowledge of the computation graph structures. 
We present the implementation details of \Echo{} in Section~\ref{sec:impl_details}.




%% file: sections/5-Implementation_Details.tex
\section{Implementation Details}
\label{sec:impl_details}

We integrate the \Echo{} compiler pass into the NNVM~\cite{nnvm} graph compilation workflow, which is used as the computation graph compiler middle-end for the MXNet~\cite{mxnet} machine learning framework. Its compilation workflow is shown in Figure~\ref{fig:nnvm_workflow}.

\begin{figure}[ht]
    \centering
    \small
    \begin{tabular}{ll}
        \toprule
        \textbf{Pass Name} & \textbf{Pass Description} \\
        \midrule
        \multirow{2}{*}{\textbf{Gradient}}
            & Recomputation (Optional) \\
            & Gradient Node Insertion \\
        \textbf{InferShape \& Type}
            & Shape and Data Type Inference \\
        \textbf{PlanMemory}
            & Device Memory Allocation Planning \\
        \bottomrule
    \end{tabular}
{\small
\begin{alignat*}{3}
& \mathbf{Gradient}
    && \rightarrow \mathbf{InferShape\ \&\ Type}
    && \rightarrow \mathbf{PlanMemory}
\end{alignat*}
}
    \vspace{-10pt}
    \captionsetup{justification=centering}
    \caption{MXNet NNVM Pass Workflow (in Sequence)}
    \label{fig:nnvm_workflow}
\end{figure}
\vspace{-3pt}

\subsection{Footprint Reduction Estimation}
\label{subsec:impl_details:footprint}

We observe that with the current graph compilation workflow, it is impossible to accurately estimate footprint reduction, because critical information such as the shape and data type of each tensor edge is only available after the recomputation algorithm has been executed. We know from the example in Section~\ref{subsec:key_ideas:footprint} (Figures~\ref{fig:add_and_tanh}~and~\ref{fig:mlp_mirror_local}) that this information is required to accurately estimate footprint reduction. 
We therefore start by changing the compilation workflow in Figure~\ref{fig:nnvm_workflow} to Figure~\ref{fig:adjusted_nnvm_workflow}, so that more information is available by the time \Echo{} is executed.

\begin{figure}[ht]
{\small
\vspace{-\baselineskip}
\begin{alignat*}{4}
& \mathbf{Gradient}
    && \rightarrow \mathbf{InferShape\ \&\ Type}
       \rightarrow \mathbf{EdgeUseRef} \\
&   && \rightarrow \text{\textbf{\emph{Echo}}} 
       \rightarrow \mathbf{DeadNodeElimination} \\
&   && \rightarrow \mathbf{InferShape\ \& \ Type} 
       \rightarrow \mathbf{AllocateMemory}
\end{alignat*}
}
    \vspace{-15pt}
    \captionsetup{justification=centering}
    \caption{Adjusted NNVM Pass Workflow \\ 
        (The first \emph{Gradient} pass does not perform recomputation)}
    \label{fig:adjusted_nnvm_workflow}
\end{figure}

With the new workflow, allocation-relevant information (e.g., shape and data type) is now available prior to \Echo{}, which gives it the opportunity to accurately estimate the footprint reduction brought by recomputation using a bidirectional dataflow analysis. \CHANGE{Algorithm~\ref{algo:echo-analysis_workflow} shows \Echo{}'s high-level workflow. 
\Echo{} first performs a backward pass (line 6-10) to obtain all the possible targets for recomputation, and partitions the entire graph at the compute-heavy layers to avoid significant overhead in case these layers are included in recomputation. It then performs a forward pass (line 12-28) to remove recomputations that do not reduce memory footprint. The forward pass reduces the recomputation runtime overhead, and also guarantees that the total memory consumption after recomputation is performed will never increase compared with the baseline.}

\newcommand\CommentFont[1]{\rmfamily\textcolor{blue}{#1}}
\SetCommentSty{CommentFont}
\LinesNumbered

\begin{algorithm}[!t]\footnotesize\CHANGE{
    \caption{\Echo{}'s High-Level Workflow}
    \label{algo:echo-analysis_workflow}
    
    \SetKw{Continue}{continue}
    \SetKwInOut{Input}{Input}
    \Input{Computation Graph \(G\)}
    
    worklist \(H=[G.\mathrm{OutputNodes}]\)\;
    \While {\(!H.\operatorname{empty}()\)} {
        \(h=H.\operatorname{pop}()\)\;
        \lIf {\(h\in G.\mathrm{InputPlaceholders}\)} {
            \Continue}
        subgraph \(S=\{h\}\), worklist \(W=[h.\mathrm{InputNodes}]\)\;
        \tcc{1. Backward: Expand the subgraph \(S\) backward until blocked by compute-heavy layers}
        \While {\(!W.\operatorname{empty}()\)} {
            \(w = W.\operatorname{pop}()\)\;
            \lIf {\(w\in G.\mathrm{InputPlaceholders}\)} { \Continue }
            \lIf {\(w.\mathrm{op}\in \mathrm{ComputeHeavyOps}\)} {
                \newline\(H.\operatorname{append}(w)\); \Continue}
            \(S.\operatorname{insert}(w)\);
            \(W.\operatorname{append}(w.\mathrm{InputNodes})\)\;
        }
        create the recomputation path of \(S\) (Figure~\ref{subfig:backward_example:result})\;
        \tcc{2. Forward: Traverse through the subgraph \(S\) in topological order and estimate the footprint effect \(\forall\)recomputation}
        \For {\(s\in S.\operatorname{TopologicalOrderView}\)} {
            \If {\(s.\mathrm{op}\in \mathrm{ComputeHeavyOps}\)} {
                create a dummy gradient node \(gs'\) and link it to \(s\) via the gradient function \(\operatorname{GradFunc}(s.\mathrm{op})\)\;
                \For {\(e\in s.\mathrm{InputEdges}\) that is feature maps (i.e., \(\in gs'.\mathrm{InputEdges}\))} {
                    modify the graph as in Figure~\ref{subfig:recompute_fc:low_overhead}\;
                }
                \Continue\;
            }
            \lIf {\(s.\mathrm{op}\in \mathrm{BinarizableOps}\)} {
                insert encode and decode subroutines between \(s\) and its gradient node \(gs\); \Continue}
            edges set \(\mathrm{AllocEdges}=\mathrm{RelEdges}=\{\}\)\;
            \For {\(e\in s.\mathrm{InputEdges}\)} {
                \For {\(n\in S\cup S.\mathrm{GradientGraph}\) that \newline
                    references \(e\) as input} {
                    \uIf {\(n\in S\)} {
                        \(\mathrm{AllocEdges}.\operatorname{insert}(n.\mathrm{OutputEdges})\)\;
                        \(\mathrm{RelEdges}.\operatorname{insert}(n.\mathrm{InputEdges})\)\;
                    } \lElse {\(\mathrm{AllocEdges}.\operatorname{insert}(e)\)}
                }
            }
            \lFor {\(e\in\mathrm{AllocEdges}\)} {
                \(\mathrm{Alloc}\mathrel{+}=e.\mathrm{Size}\)}
            \lFor {\(e\in\mathrm{RelEdges}\)} {
                \(\mathrm{Rel}\mathrel{+}=e.\mathrm{Size}\)}
            \lIf {\(\mathrm{Rel}\ge\mathrm{Alloc}\)} {
                remove \(s\) from the mirror path (Figure~\ref{subfig:forward_example:1} and Figure~\ref{subfig:forward_example:2})}
        }
    }
}\end{algorithm}

\begin{figure}[ht]
    \centering
    \captionsetup{justification=centering}
    \begin{subfigure}[t]{0.3\linewidth}
        \centering
        \includegraphics[width=54pt]{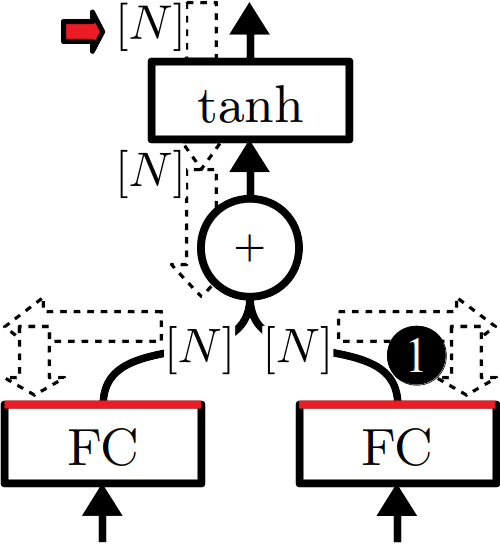}
        \vspace{-5pt}
        \caption{Backward \newline Analysis Flow}
        \label{subfig:backward_example:1}
    \end{subfigure}
    \begin{subfigure}[t]{0.3\linewidth}
        \centering
        \includegraphics[width=54pt]{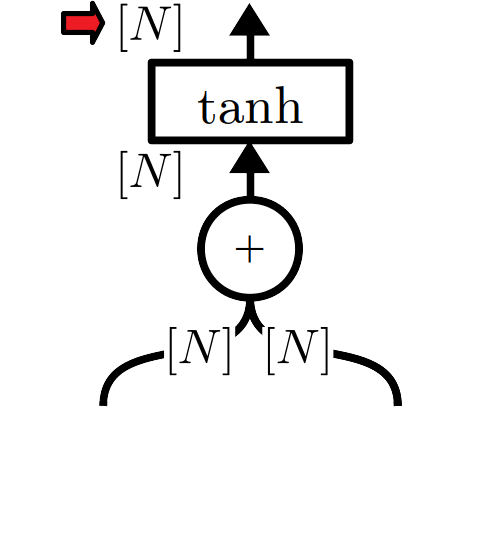}
        \vspace{-5pt}
        \caption{Partitioned \newline Subgraph}
        \label{subfig:backward_example:2}
    \end{subfigure}
    \begin{subfigure}[t]{0.3\linewidth}
        \centering
        \includegraphics[width=54pt]{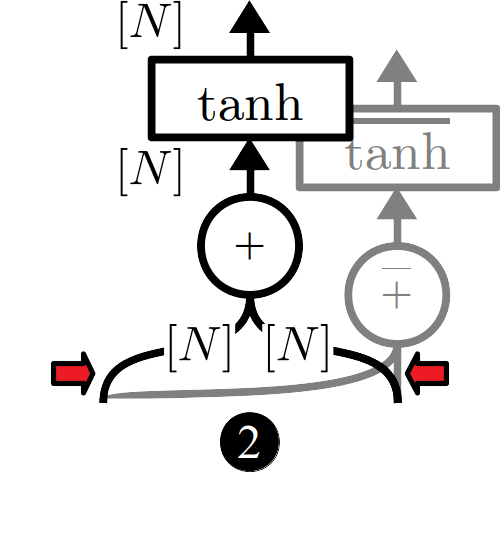}
        \vspace{-5pt}
        \caption{Recomputation Path}
        \label{subfig:backward_example:result}
    \end{subfigure}
    \captionsetup{justification=centering}
    \caption{Backward Analysis Example}
    \label{fig:backward_example}
\end{figure}

Figure~\ref{fig:backward_example} illustrates an example similar to Figure~\ref{fig:add_and_tanh}, with two fully-connected layers added before the elementwise-add operator. The backward pass, shown in Figure~\ref{subfig:backward_example:1}, starts from the top (i.e., output) of the graph, propagates backward along the dashed edges, and then stops at the fully-connected layers (\darkcircled{1}), because these two layers do not belong to the possible targets for recomputation as being too compute-heavy (see the runtime profile in Section~\ref{subsec:motivation:runtime}, Figure~\ref{fig:sockeye-runtime_profile}). 
 
After the backward pass, \Echo{} takes the nodes and edges that the backward pass has processed (Figure~\ref{subfig:backward_example:2}), which form a partitioned subgraph of the original computation graph, and assumes that all the operator nodes within this subgraph can be recomputed. This is illustrated in Figure~\ref{subfig:backward_example:result} as a graph shown in gray that mirrors the subgraph. The mirrored graph is the \emph{recomputation path}, similar to the gray segments in Figures~\ref{fig:4_tanh_nodes} and \ref{fig:add_and_tanh}. Due to the recomputation path, the feature maps that are originally at the output edge of \(\tanh\) in Figure~\ref{subfig:backward_example:1}, are now placed at the beginning of the recomputation path (\darkcircled{2}), as we have shown in Section~\ref{subsec:motivation:backward_mirroring}, Figure~\ref{fig:4_tanh_nodes}. 


After the recomputation path has been formed, a followup forward pass removes recomputations that cannot reduce memory footprint. 
A node can be removed from the recomputation path if, by removing that node, the storage released from its inputs is greater than or equal to that allocated for its outputs.

\begin{figure}[ht]
    \centering
    \captionsetup{justification=centering}
    \begin{subfigure}[t]{0.3\linewidth}
        \centering
        \includegraphics[height=60pt]{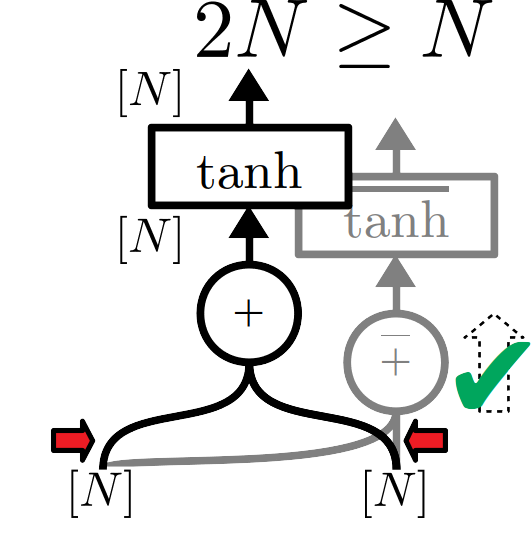}
        \vspace{-7pt}
        \caption{Forward on \newline Elementwise Add}
        \label{subfig:forward_example:1}
    \end{subfigure}
    \begin{subfigure}[t]{0.3\linewidth}
        \centering
        \includegraphics[height=60pt]{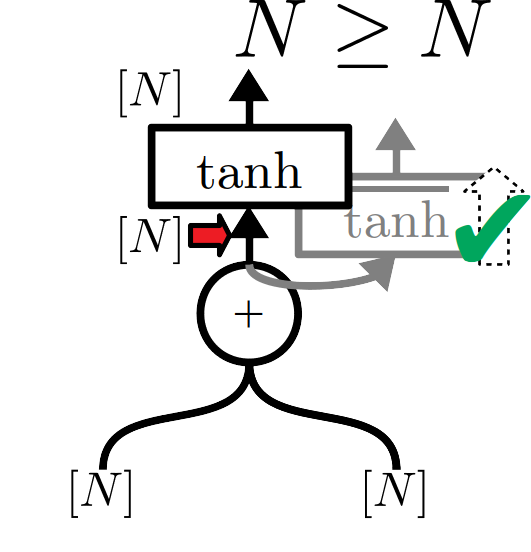}
        \vspace{-7pt}
        \caption{Forward on \newline \(\tanh\) Activation}
        \label{subfig:forward_example:2}
    \end{subfigure}
    \begin{subfigure}[t]{0.3\linewidth}
        \centering
        \includegraphics[height=60pt]{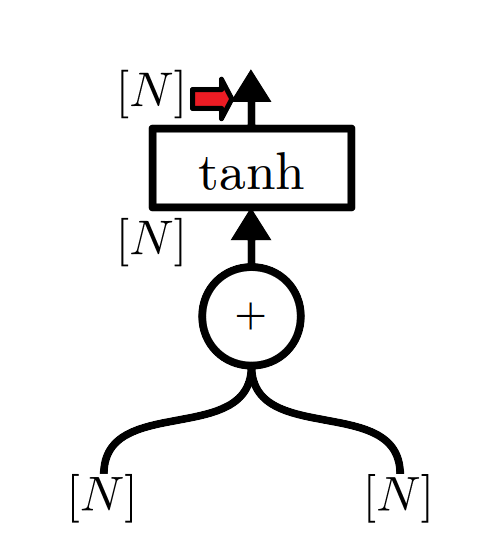}
        \vspace{-7pt}
        \caption{Final Graph}
        \label{subfig:forward_example:final_graph}
    \end{subfigure}
    \caption{Forward Analysis Example (Figure~\ref{fig:backward_example} Continued)}
    \label{fig:forward_example}
\end{figure}

Based on the previous example, we consider to remove the \(+\) and \(\tanh\) nodes from the recomputation path in succession. 
Figure~\ref{subfig:forward_example:1} shows that the removal of \(+\) is successful, because its inputs size (\(N+N\)) is greater than its output (\(N\)). Similarly, Figure~\ref{subfig:forward_example:2} shows that the removal of \(\tanh\) is also successful, because its input size (\(N\)) is equal to its output size (\(N\)).
After the removals, the final graph (Figure~\ref{subfig:forward_example:final_graph}) does not have any recomputation. 
This indicates that recomputation is unable to reduce memory footprint in this specific example (as we have seen in Section~\ref{subsec:key_ideas:footprint}), but \Echo{} is still able to preserve both the same performance and memory footprint as the baseline. This is in stark contrast to prior works \cite{sublinear, memory_effi-bptt, hessian-free} that introduce unnecessary recomputation which doubles the feature maps storage (shown in Section~\ref{subsec:key_ideas:footprint}, Figure~\ref{subfig:add_and_tanh:mirror}).


After the forward pass, the backward pass resumes at the inputs of the fully-connected layers in Figure~\ref{subfig:backward_example:1} for the next partitioned subgraph. This backward-forward loop continues until the inputs of the whole graph are reached. As one might notice, since the next subgraph continues at the places where the previous subgraph stops, each subgraph is \emph{disjoint}. \CHANGE{In practice, for the wide set of models in our evaluation (Section~\ref{sec:evaluation}), the number of subgraphs ranges from \(100\) to \(2000\), with each one including no more than 10 operator nodes. These properties (disjoint and small) keep the runtime complexity of Algorithm~\ref{algo:echo-analysis_workflow} reasonable (less than \(300\ \mathrm{ms}\) for the models in Section~\ref{sec:evaluation} in the hardware environment listed in Section~\ref{subsec:evaluation:methodology}).}


As we have discussed in the example in Section~\ref{subsec:key_ideas:footprint} (Figure~\ref{fig:mlp_mirror_local}), the comparison between the storage released and allocated requires more than just the comparison between input versus output sizes that only reflects allocations that are \emph{local} to each operator. For accurate footprint reduction estimation, \Echo{} considers the \emph{global effect} (i.e., \emph{reuse}) of storage allocations within each subgraph \emph{independently}, because each subgraph is disjoint and hence there is no reuse across subgraphs. \Echo{} abstracts the \emph{reuse} using the \emph{use references} on each tensor edge, as it represents the number of times a particular tensor (and hence the storage allocated to that tensor) is \emph{reused}. The tensor edge use references come from the \emph{EdgeUseRef} pass (see Figure~\ref{fig:adjusted_nnvm_workflow}), where the compiler traverses through the whole computation graph and, for each tensor, records the number of times it is referenced by different operators. 

\begin{figure}[ht]
    \centering
    \begin{subfigure}[t]{0.457\linewidth}
        \centering
        \includegraphics[width=\linewidth]{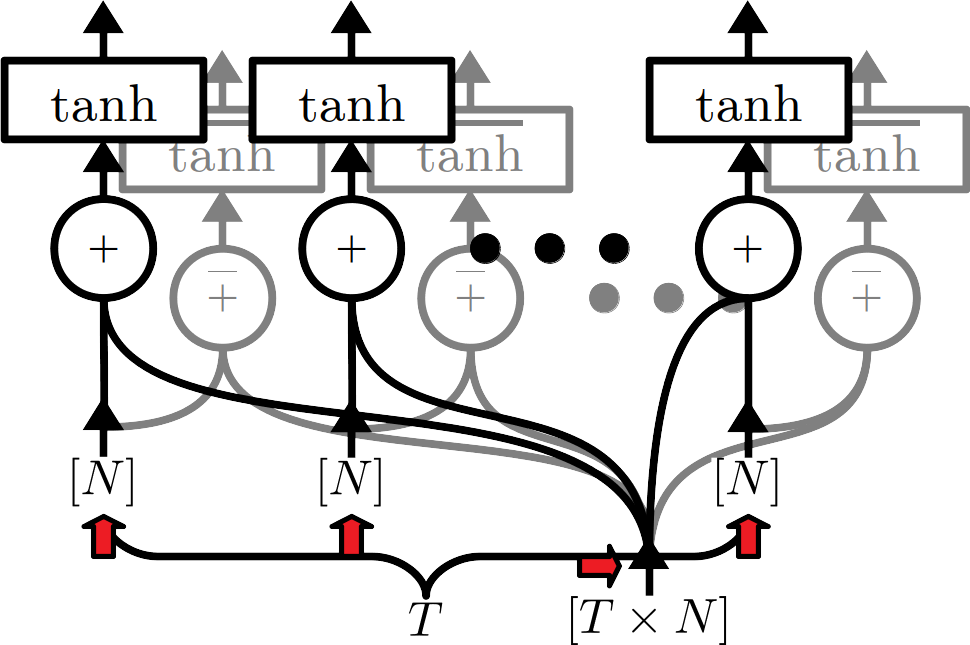}
        \vspace{-15pt}
        \caption{Recomputation Path}
        \label{subfig:storage_ref:backward}
    \end{subfigure}
    \begin{subfigure}[t]{0.523\linewidth}
        \centering
        \includegraphics[width=\linewidth]{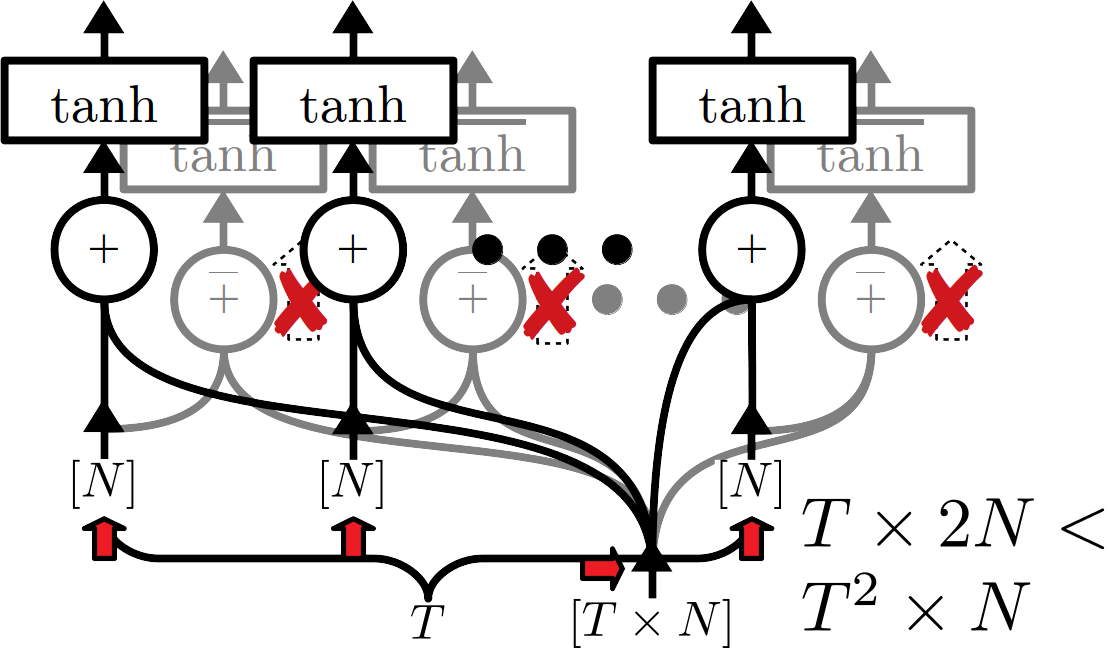}
        \vspace{-15pt}
        \caption{Final Graph}
        \label{subfig:storage_ref:forward}
    \end{subfigure}
    \captionsetup{justification=centering}
    \caption{Analysis Example based on Figure~\ref{fig:mlp_mirror_local}}
    \label{fig:storage_ref}
\end{figure}

Figure~\ref{fig:storage_ref} shows how \Echo{} leverages the use references information in its analysis. Let us revisit the example illustrated in Section~\ref{subsec:key_ideas:footprint} (Figure~\ref{fig:mlp_mirror_local}), where we have \(T\) tensors of shape \([N]\) broadcast-added with the same tensor of shape \([T\times N]\) and then \(\tanh\)-activated. Although the backward pass in this example (shown in Figure~\ref{subfig:storage_ref:backward}) is similar to that in Figure~\ref{subfig:backward_example:result}, where all operators are listed as recomputed, the forward pass is different. This is because the tensor \([T\times N]\) is used by \(T\) different operators, and trying to remove any of those operators on the recomputation path will cause all the other operators to be removed as well, since they all need the tensor \([T\times N]\) to be stashed as feature maps to do recomputation. This is illustrated as \(T\) parallel arrows on the recomputation path in Figure~\ref{subfig:storage_ref:backward}. \Echo{} therefore needs to compare the storage allocated for \emph{all} \(T\) operators' inputs with their outputs. It then notices that, since the storage allocated for \([T\times N]\) is shared by all the \(T\) operators, the total inputs storage size (\(T\times N+T\times N=T\times 2N\)) is smaller than the outputs (\(T\times T\times N=T^2\times N\)) when \(T\) is large enough (usually \(T\) is in the range of \(50\sim 100\) \cite{nmt, sockeye}). \Echo{} therefore stops trimming the recomputation path right at the beginning, leaving all operators marked as recomputed. Indeed, the total feature maps storage in the final graph is \(T\times 2N\), which is an order of magnitude smaller than storage required in the baseline (\(T^2\times N\)).

We conclude that, compared with prior works \cite{sublinear, memory_effi-bptt, hessian-free} that na\"ively ignore compute-heavy operators as potential targets for recomputation. \Echo{} uses rigorous dataflow analysis to avoid pathological cases where recomputation is not needed while preserving those where it is beneficial. This explains why \emph{\Echo{} keeps overhead minimal and never increases the memory footprint}, as our results in Section~\ref{subsubsec:evaluation:generality:ds2} will show.

\subsection{Runtime Overhead Estimation}
\label{subsec:impl_details:dead_node}

\begin{figure}[ht]
    \centering
    \captionsetup{justification=centering}
    \begin{subfigure}[t]{0.33\linewidth}
        \centering
        \includegraphics[width=\linewidth]{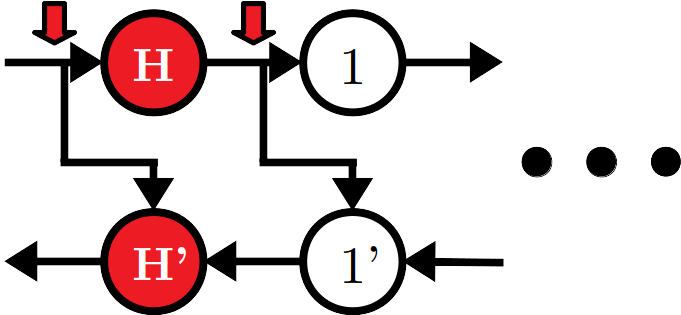}
        \vspace{-15pt}
        \caption{Baseline}
    \end{subfigure}
    \hspace{0.05\linewidth}
    \begin{subfigure}[t]{0.33\linewidth}
        \centering
        \includegraphics[width=\linewidth]{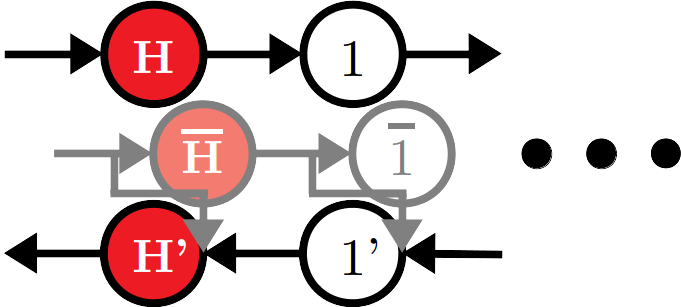}
        \captionsetup{width=1.1\linewidth}
        \vspace{-15pt}
        \caption{High Recomputation \newline Overhead}
        \label{subfig:recompute_fc:costly}
    \end{subfigure}
    \\
    \captionsetup{justification=centering}
    \begin{subfigure}[b]{0.33\linewidth}
        \centering
        \includegraphics[width=\linewidth]{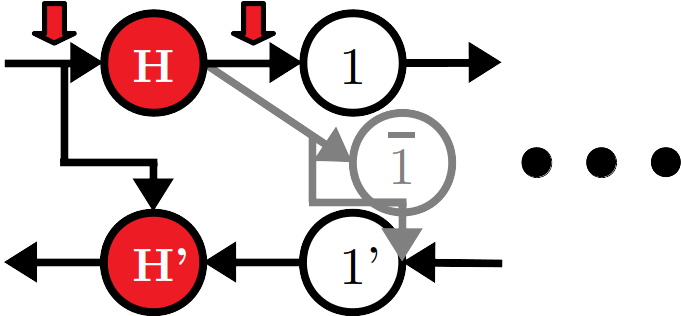}
        \captionsetup{width=1.1\linewidth}
        \vspace{-15pt}
        \caption{Prior Works \cite{sublinear} \newline No Footprint Reduction}
        \label{subfig:recompute_fc:no_reduction}
    \end{subfigure}
    \hspace{0.05\linewidth}
    \begin{subfigure}[b]{0.329\linewidth}
        \centering
        \includegraphics[width=\linewidth]{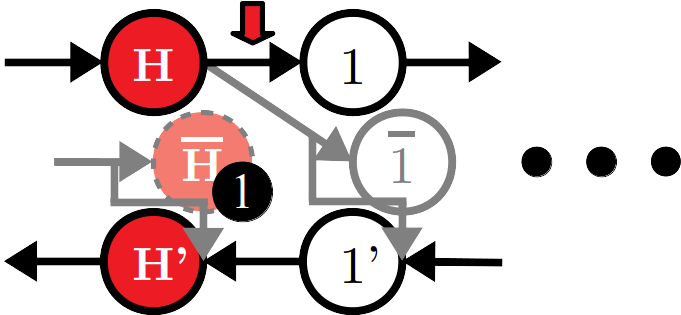}
        \vspace{-15pt}
        \caption{Low-Overhead \newline Footprint Reduction}
        \label{subfig:recompute_fc:low_overhead}
    \end{subfigure}
    \captionsetup{justification=justified}
    \caption{Recomputation Strategies for \CHANGE{Compute-Heavy Layers (shown as Red \(H\)'s) whose Gradients only need \(H\)'s Inputs}}
    \label{fig:recompute_fc}
\end{figure}

As in Section~\ref{subsec:key_ideas:runtime}, the gradients of the fully-connected layers only need their inputs. Layers such as convolutions and batched dot product also have similar property. Prior works \cite{sublinear} simply skip all the compute-heavy layers (Figure~\ref{subfig:recompute_fc:no_reduction}). 
In contrast, \Echo{} deems these layers as potentially good targets for recomputation by performing non-conservative runtime overhead estimation \CHANGE{(Algorithm~\ref{algo:echo-analysis_workflow} line 13-17)}. \Echo{} starts by inferring the data dependencies of the gradient operator that are specific to different types of layers. It does so by (i) creating a dummy gradient operator node, (ii) applying the gradient function to the (gradient, forward) operator tuple, and (iii) analyzing the data dependencies of the gradient operator on the forward operator. 
If the dependencies do not include the outputs of the forward operator, this implies that recomputation in this case does not require to recompute the operator itself, and hence its runtime is \emph{not} added to the total runtime overhead when recomputing feature maps. If this is the case, \Echo{} creates a ``dead'' node on the recomputation path (\darkcircled{1}) that forwards the output edge of the previous recomputation node to the gradient node. However, the node itself is disconnected from the next node on the recomputation path. This implies that the node's outputs are never referenced by any other nodes, making the node effectively \emph{dead} and thereby avoiding any unnecessary recomputation. Such an approach releases the feature maps on the input edges of the compute-heavy nodes but does not lead to situations with huge runtime overhead (as in Figure~\ref{subfig:recompute_fc:costly}), and can therefore further reduce the GPU memory footprint.

Although the newly inserted dead nodes are never referenced, they still remain as part of the computation graph and hence require storage allocations that are not necessary. We introduce the \emph{DeadNodeElimination} pass that properly cleans up those auxiliary nodes after \Echo{} finishes (see Figure~\ref{fig:adjusted_nnvm_workflow}).

In addition to leveraging the layer specific gradient dependencies, \Echo{} also uses layer specific encoding to binarize~\cite{gist} the feature maps of the \(\operatorname{dropout}\) layers \cite{dropout} \CHANGE{(Algorithm~\ref{algo:echo-analysis_workflow} line~18)}. 
It encodes the dropout feature maps to \(1\)-bit in the forward pass and decodes them back to \(32\)-bit in the backward pass. 
Such layer specific optimizations give more footprint reduction with small runtime overhead. 

In summary, we have demonstrated how \Echo{} benefits from layer specific information and uses non-conservative runtime overhead estimation to reduce the memory footprint with low runtime overhead, 
as our results in Section~\ref{sec:evaluation} will show.

%% file: sections/6-Evaluation.tex
\section{Evaluation}
\label{sec:evaluation}

\subsection{Methodology}
\label{subsec:evaluation:methodology}

\textbf{Infrastructure.}
Our major compute platform is a single machine with \(32\)-core AMD EPYC 7371 CPU~\cite{amd-epyc-7371} (with \(128\ \mathtt{GB}\) DDR4 \cite{ddr4} memory) and \(4\) NVIDIA RTX 2080 Ti GPUs~\cite{2080-ti} (Turing architecture \cite{turing} with \(11\ \mathtt{GB}\) GDDR6 memory \cite{gddr6}) connected via PCIe v3 \cite{pcie3_0}, installed with CUDA 10.0 \cite{cuda-10}, cuDNN 7.6.3 \cite{cudnn-7_6_3}, and MXNet v0.12.1 \cite{mxnet-0121}. 


\textbf{Applications.} 
We evaluate \Echo{} by training the Sockeye~\cite{sockeye} NMT toolkit on the IWSLT15 English-Vietnamese (small)~\cite{iwslt15} and WMT16 English-German (large)~\cite{wmt16} datasets, using the hyperparameters from \citet{tbd} for the single-GPU experiments on the small dataset and \citet{sockeye} for the multi-GPU experiments on the large dataset. 

We also demonstrate results on DeepSpeech2 \cite{deepspeech2}, the state-of-the-art (SOTA) speech recognition model \cite{mlperf, tbd}, using the LibriSpeech dataset \cite{librispeech}; on Transformer \cite{transformer} (SOTA machine translation model \cite{mlperf, tbd}) using the WMT16 English-German dataset \cite{wmt16}; and on ResNet-152 (SOTA image classification model \cite{mlperf, tbd}) using the ImageNet dataset \cite{imagenet}.

\textbf{Baselines.}
We compare our fully automated and transparent approach, \Echo{}, with two baselines: the baseline system without recomputation, which we refer to as \Baseline{}, and the state-of-the-art implementation of recomputation by \citet{sublinear}, which we refer to as \Mirror{}. \CHANGE{Although recomputation is also implemented in other frameworks (e.g., TensorFlow \cite{openai-grad_ckpt}), those implementations follow the prior work \cite{sublinear} with the limitations we previously described. These limitations have been flagged as causing memory issues (e.g., increasing the GPU memory footprint instead of decreasing it \cite{openai-grad_ckpt-issue}), making the usability of these implementations questionable.}

For the experiments on NMT \cite{nmt, gnmt}, we use the superscript (\(\dagger\)) to denote our hand-tuned implementation. We leverage the information provided by \Echo{} to pinpoint the places where recomputation is beneficial and manually fuse the operators that are on the recomputation path into a single operator (i.e., Node \#\(\overline{1}\sim \overline{4}\) in Figure~\ref{subfig:4_tanh_nodes:mirror} are fused together as a single CUDA kernel). As we discussed in Section~\ref{subsec:key_ideas:compiler}, this requires non-trivial effort and expertise in CUDA programming, machine learning algorithms, and framework system integration, but it shows potential benefits that recomputation can provide.

\textbf{Metrics.}
We show the (1) GPU memory consumption and (2) throughput as the key metrics. We also show the training curves in the NMT experiments, and power/energy consumption on the GPUs in the multi-GPU experiments. The training curves are expanded using the CPU wall clock time. Those curves use BLEU score \cite{bleu} to quantify the machine translation quality, where a higher BLEU score means better translation quality and a BLEU score that is greater than \(20\) is considered strong \cite{sockeye, tf-nmt, opennmt, bleu}.
As training progresses, we periodically query the memory and power usage of the GPUs using the \emph{nvidia-smi} tool \cite{nvidia-smi} and approximate the GPU energy consumption as power over time. The throughput is reported as the average of the throughput numbers given by the MXNet speedometer~\cite{mxnet-speedometer}, which measures throughput by dividing the number of training samples by the CPU wall clock time.

\subsection{Machine Translation Results with NMT Model}
\label{subsec:evaluation:nmt}


\subsubsection{English-Vietnamese}
\label{subsubsec:evaluation:nmt:iwslt15}
\begin{figure}[!t]
    \centering
    \includegraphics[height=24pt]{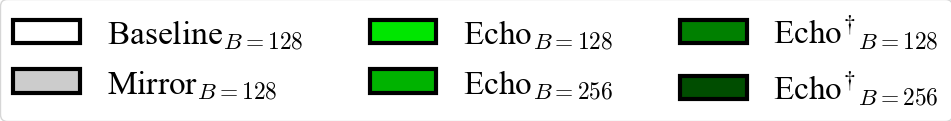}
    \includegraphics[height=12pt]{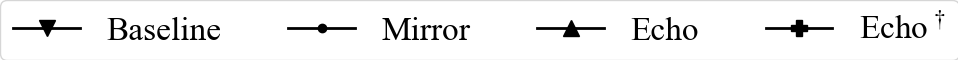}
    \begin{minipage}[b]{0.49\linewidth}
        \begin{subfigure}[b]{\linewidth}
            \includegraphics[width=\linewidth]{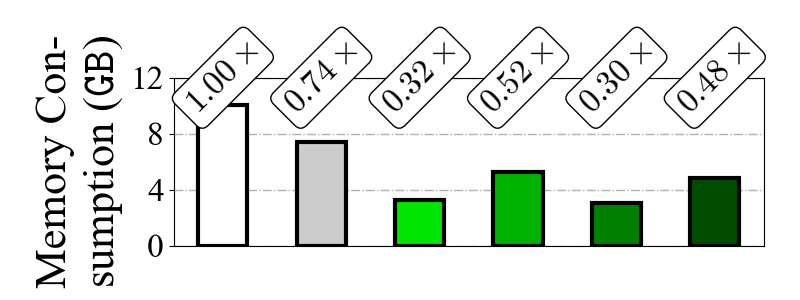}
            \vspace{-15pt}
            \caption{GPU Memory Consumption}
            \label{subfig:iwslt15:memory}
        \end{subfigure}
        \begin{subfigure}[b]{\linewidth}
            \includegraphics[width=\linewidth]{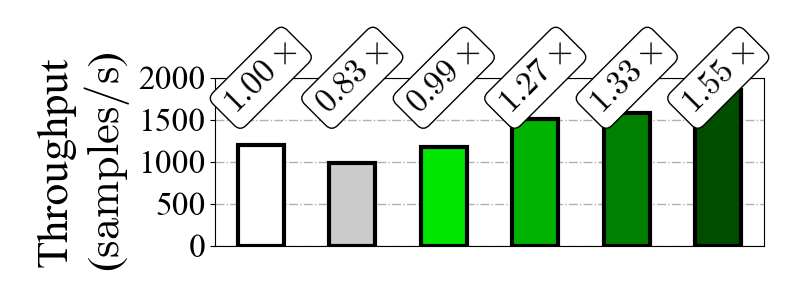}
            \vspace{-15pt}
            \caption{Throughput}
            \label{subfig:iwslt15:thruput}
        \end{subfigure}
    \end{minipage}
    \begin{minipage}[b]{0.49\linewidth}
        \begin{subfigure}[b]{\linewidth}
            \includegraphics[width=\linewidth]{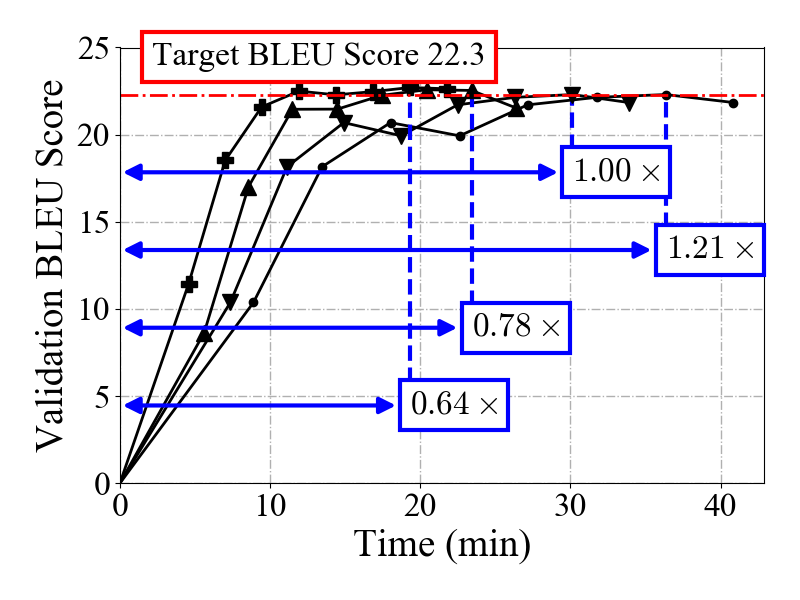}
            \vspace{-12pt}
            \caption{Validation Curve BLEU Score}
            \label{subfig:iwslt15:bleu}
        \end{subfigure}
    \end{minipage}
    \caption{(a) GPU Memory Consumption, (b) Throughput, and (c) Validation Curve BLEU Score compared between \Baseline{}, \Mirror{}, and \Echo{}\(^{(\dagger)}\) on English-Vietnamese NMT} 
    \label{fig:iwslt15:memory_thruput_bleu}
\end{figure}

Figure~\ref{subfig:iwslt15:memory} and \ref{subfig:iwslt15:thruput} illustrate the comparison of the GPU memory usage and the training throughput under different batch sizes on the English-Vietnamese translation task (the subscript \(B\) denotes the batch size used). We observe that \Echo{} can achieve \NMTFootprintReductionRatio{} footprint reduction ratio over \Baseline{} and \(2.31\times\) over \Mirror{} under the same batch size of \(128\) \CHANGE{(\(35.4\%\) of the reduction comes from more aggressive recomputation due to non-conservative runtime overhead estimation (Section~\ref{subsec:key_ideas:runtime})}. We also observe that \Echo{} only has \(1\%\) runtime overhead, which is \(18\times\) less than \Mirror{}.

Because \Baseline{}\(\smash{_{B=128}}\) and \Mirror{}\(\smash{_{B=128}}\) consume around \(10.0\ \mathtt{GB}\) and \(7.4\ \mathtt{GB}\) of GPU memory respectively when the batch size is \(128\), their batch size can no longer be doubled (otherwise the GPU will run out of memory). The situation is different for \Echo{}. Since it only consumes around \(3.0\ \mathtt{GB}\) of memory, \Echo{}'s training batch size can be further increased to \(256\), producing the training curve \Echo{}\(\smash{_{B=256}}\) in Figure~\ref{subfig:iwslt15:bleu}. By training the NMT model with larger batches, we increase the throughput by \(1.27\times\) and converge to the same validation BLEU score \IWSLTSpeedup{} faster than the baseline. \CHANGE{The reason why the achieved speedup \IWSLTSpeedup{} is smaller than those shown in Figure~\ref{subfig:sockeye} is two-fold: (1) Similar to the ResNet model~\cite{resnet} (Figure~\ref{subfig:resnet_50}), throughput can saturate at large batch size, as the compute resource utilization increases with the batch size. 
(2) The recomputation overhead lessens the performance benefits of having a larger batch size. However, such negative impact can be mitigated by the hand-tuned implementation \Echo{}\(^{\dagger}\).}
As one might notice, at the same batch size of \(128\), \Echo{}\(^{\dagger}\) (manually optimized version of \Echo{}) reduces the GPU memory footprint by \(3\times\) while increasing, rather than decreasing, the throughput by \(33\%\). The reason for the increase is because the overhead of the recomputation is so small that it is outweighed by the benefits of kernel fusion. Such benefits include reduction in (i) the \lstinline{cudaLaunch} overhead and (ii) the number of GPU memory accesses \cite{kernel-fusion}. As Figure~\ref{subfig:iwslt15:bleu} illustrates, after doubling the batch size, \Echo{}\(^\dagger\) improves the speed of convergence further by \(1.56\times\) compared with the baseline.

\subsubsection{English-German}
\label{subsubsec:evaluation:nmt:multi_gpu}

Multi-GPU training is a common way to reduce the training time \cite{gnmt, deepspeech}, however, in multi-GPU training, communication can potentially become a bottleneck. Moreover, power and energy is now a primary concern as GPU cards such as RTX 2080 Ti is known for being power-hungry (having a TDP at around \(250\ \mathrm{W}\)) \cite{2080-ti}.

\begin{figure*}[!t]
    \begin{minipage}[t]{0.49\linewidth}
        \centering
        \includegraphics[height=14pt]{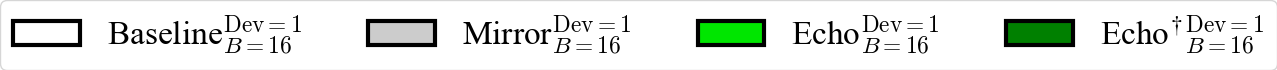}
        \begin{subfigure}[b]{0.49\linewidth}
            \includegraphics[width=\linewidth]{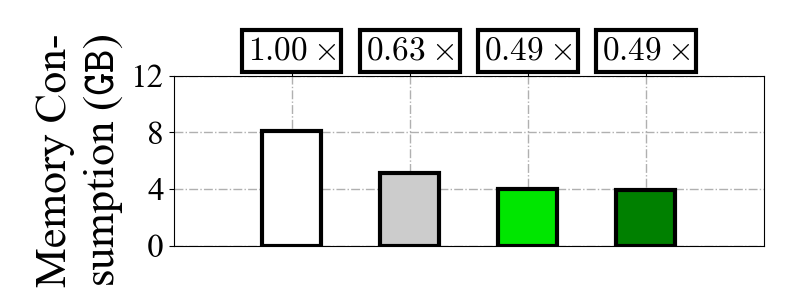}
            \vspace{-15pt}
            \caption{GPU Memory Consumption}
            \label{subfig:wmt16:single_gpu:memory}
        \end{subfigure}
        \begin{subfigure}[b]{0.49\linewidth}
            \includegraphics[width=\linewidth]{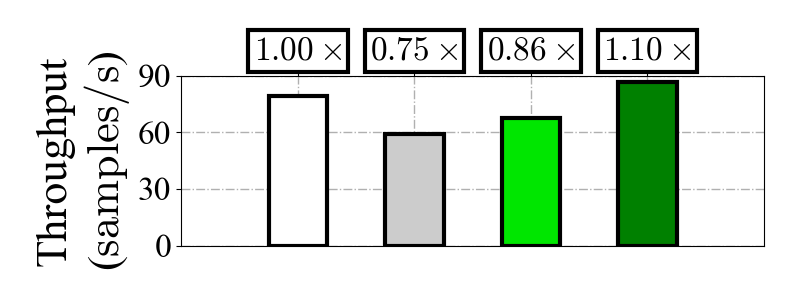}
            \vspace{-15pt}
            \caption{Throughput}
            \label{subfig:wmt16:single_gpu:throughput}
        \end{subfigure}
    \end{minipage}
    \begin{minipage}[t]{0.49\linewidth}
        \centering
        \includegraphics[height=24pt]{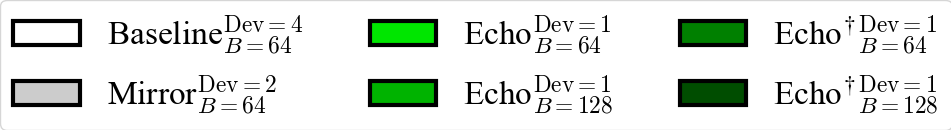} 
        \begin{subfigure}[b]{0.49\linewidth}
            \includegraphics[width=\linewidth]{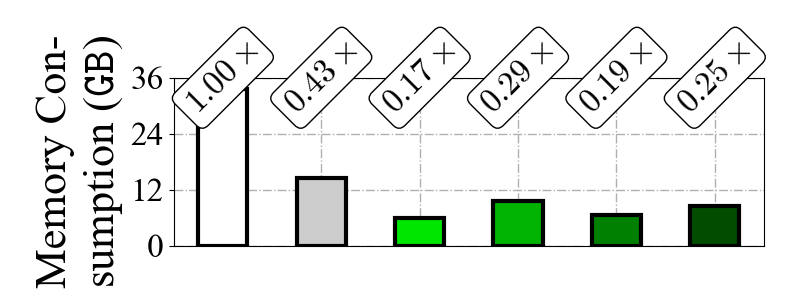}
            \vspace{-15pt}
            \caption{GPU Memory Consumption}
            \label{subfig:wmt16:multi_gpu:memory}
        \end{subfigure}
        \begin{subfigure}[b]{0.49\linewidth}
            \includegraphics[width=\linewidth]{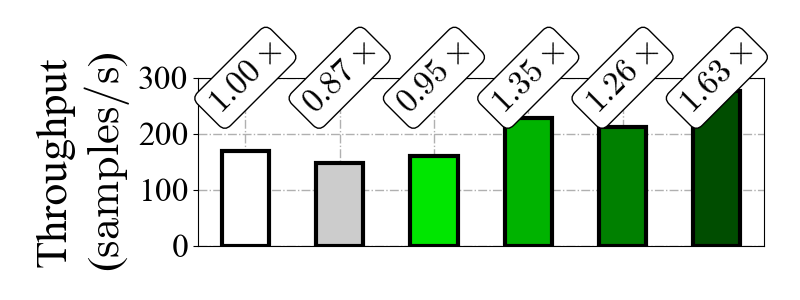}
            \vspace{-15pt}
            \caption{Throughput}
            \label{subfig:wmt16:multi_gpu:throughput}
        \end{subfigure}
    \end{minipage}
    \captionsetup{width=0.9\linewidth}
    \caption{(a, c) GPU Memory Consumption and (b, d) Throughput
            compared between \Baseline{}, \Mirror{}, and \Echo{}\(^{(\dagger)}\) on English-German NMT (a-b: Single-GPU, \(b=16\), c-d: Multi-GPU, \(B=64/128\))}
    
    \includegraphics[height=14pt]{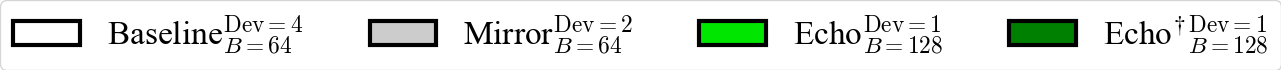}
    \includegraphics[height=14pt]{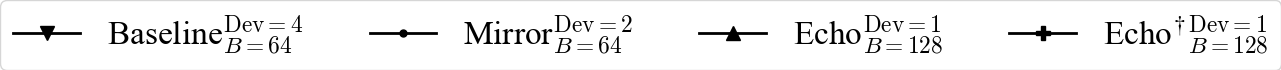}
    
    
    \begin{minipage}[t]{0.24\linewidth}
        \centering
        \includegraphics[width=\linewidth]{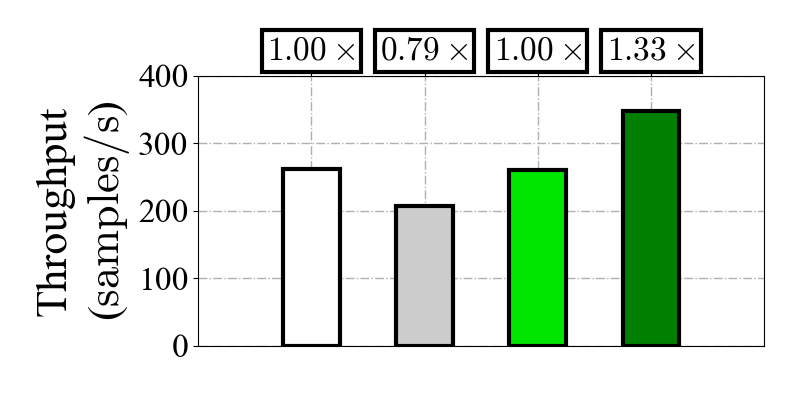}
        
        \vspace{-5pt}
        \captionsetup{width=0.95\linewidth}
        \caption{\CHANGE{Throughput Comparison on \(4\) Tesla V100s connected via NVLink}}
        \label{fig:wmt16:v100:thruput}
    \end{minipage}
    \begin{minipage}[t]{0.74\linewidth}
        \begin{subfigure}[t]{0.32\linewidth}
            \includegraphics[width=\linewidth]{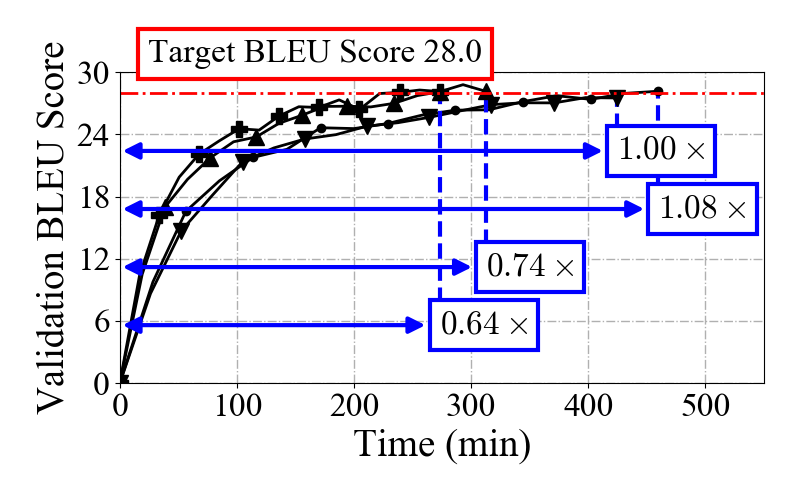}
            
            \vspace{-8pt}
            \caption{Validation Curve BLEU Score}
            \label{subfig:wmt16:multi_gpu:validation_bleu}
        \end{subfigure}
        \begin{subfigure}[t]{0.66\linewidth}
            \includegraphics[width=0.49\linewidth]{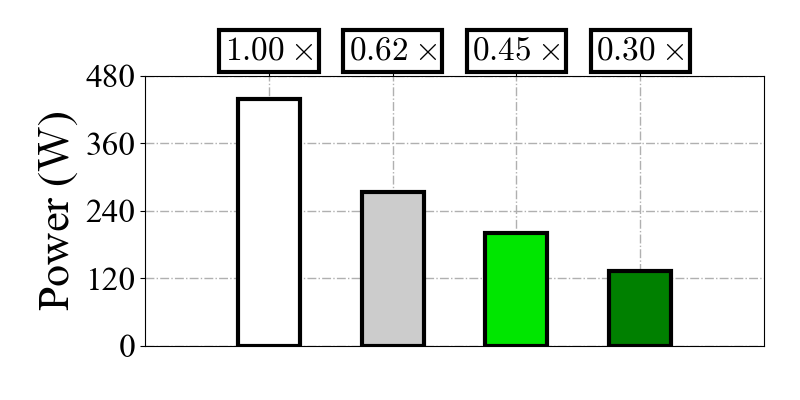}
            \includegraphics[width=0.49\linewidth]{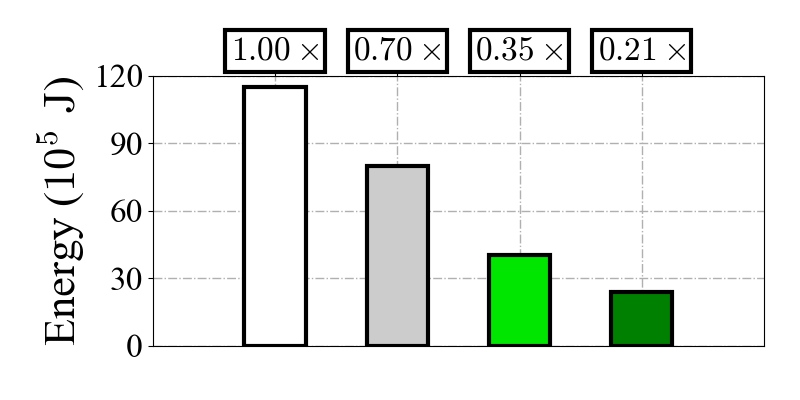}
            
            \vspace{-8pt}
            \caption{Power \& Energy Consumption}
            \label{subfig:wmt16:multi_gpu:pe}
        \end{subfigure}
        \captionsetup{width=0.95\linewidth}
        \caption{(a) Validation Curve BLEU Score and (b) Power \& Energy Consumption compared between \Baseline{}, \Mirror{}, and \Echo{}\(^{(\dagger)}\) on English-German NMT}
        \label{fig:wmt16:multi_gpu:full_training}
    \end{minipage}
\end{figure*}

Figure~\ref{subfig:wmt16:single_gpu:memory} and \ref{subfig:wmt16:multi_gpu:memory} show the memory usage under different batch sizes and device settings on the English-German translation task (the superscript \(\mathrm{Dev}\) denotes the number of GPUs used, and if multiple GPUs are used, the memory usages are aggregated up across all GPUs). We observe from Figure~\ref{subfig:wmt16:single_gpu:memory} that \Baseline{} already has a memory consumption of more than \(8\ \mathtt{GB}\) on a single GPU when the batch size is \(16\), meaning that we need to use \(4\) GPUs to train on a batch size of \(64\). However, with \Echo{} we can train on just a single GPU with a batch of \(64\), and even \(128\), as the memory consumption (\(9.6\ \mathtt{GB}\)) can still fit in the memory capacity of one RTX 2080 Ti card \cite{2080-ti}.

\CHANGE{Figure~\ref{subfig:wmt16:single_gpu:throughput} and \ref{subfig:wmt16:multi_gpu:throughput} show the throughput comparison between different implementations. We observe that although \Echo{} is \(14\%\) behind \Baseline{} when the batch size is \(16\), \Echo{} running on one GPU outperforms the baseline on four GPUs by \(1.35\times\), both using the maximum training batch size. The reason for the low scalability of the multi-GPU baseline (\(2.14\times\)) is two-fold: (1) the nature of the translation model that limits the scalability \cite{tf-nmt-multi_gpu, tf-nmt-wmt16_deen} and (2) relatively low bandwidth of the PCIe interconnect. Although NVLink-enhanced compute systems such as the ones in Amazon EC2 p3.8xlarge instance~\cite{p3_8xlarge} (\(4\) NVIDIA Tesla V100~\cite{v100} GPUs connected via NVLink~\cite{nvlink}) can be used to boost the scalability up to \(3.40\times\), such systems are significantly more expensive (as much as \(6\times\) \cite{cluster-2080_ti, cluster-v100}). Moreover, even in this hardware setup, we observe that \Echo{} running on one GPU achieves nearly the same performance with the \Baseline{} running on four GPUs, as Figure~\ref{fig:wmt16:v100:thruput} shows.}

We further observe that \Echo{} can significantly reduce the power and energy consumption on the GPUs. In Figure~\ref{subfig:wmt16:multi_gpu:validation_bleu} we show the validation BLEU score curve in one training epoch expanded by the CPU wall clock time and Figure~\ref{subfig:wmt16:multi_gpu:pe} the averaged power and accumulated energy consumption of all the GPUs. After one epoch, all implementations reach a BLEU score of \(28.0\) on the validation dataset,
however, \Echo{}\(\smash{_{B=128}^{\mathrm{Dev}=1}}\) completes the epoch \(1.35\times\) faster than \Baseline{}\(\smash{_{B=64}^{\mathrm{Dev}=4}}\). It also saves \(65\%\) of the energy consumption on the GPUs, because it requires only one GPU to train on this large batch size. 

On the other hand, although \Mirror{} can halve the number of GPUs required for training at a batch size of \(64\), it cannot further squeeze the training model onto a single GPU, let along further doubling the batch size. Hence, it is \(31\%\) behind \Echo{} in convergence speed and consumes \(2\times\) more energy on the GPUs to complete the training.

In summary, we have shown the benefit of \Echo{} on the GPU memory footprint and how to convert such benefit into faster convergence speed and/or lower GPU energy consumption. In fact, we have also implicitly shown how one could train larger and deeper model with \Echo{}, 
as the model for English-German translation has \(2\times\) more layers than that for English-Vietnamese \cite{sockeye, tbd},
but with \Echo{} they can both train properly on a single GPU under a batch size of \(128\). We conclude that \Echo{} can not only provide performance gain and energy consumption reduction, but also allow us to train deeper models with the same amount of GPU resources.


\subsection{\Echo{} Generality Across Machine Learning Models}
\label{subsec:evaluation:generality}

Since \Echo{}'s key ideas are independent of the structure of the computation graph, it is potentially applicable to any machine learning models. In this section, we evaluate \Echo{} on four state-of-the-art (SOTA) machine learning models.

\subsubsection{DeepSpeech2}\label{subsubsec:evaluation:generality:ds2}
\begin{figure}[!t]
    \centering
    \includegraphics[height=12pt]{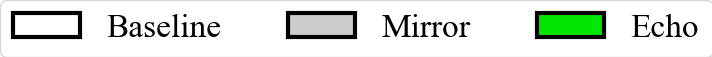}
    
    \setlength{\tabcolsep}{0pt}\renewcommand{\arraystretch}{0}
    \begin{tabular}{ccc}
        & {\footnotesize (a) GPU Memory Consumption} & 
          {\footnotesize (b) Throughput} \\
        \raisebox{-0.5\height}{\rotatebox{90}{\footnotesize (1) DS2}} &
        \raisebox{-0.5\height}{\includegraphics[width=0.46\linewidth]{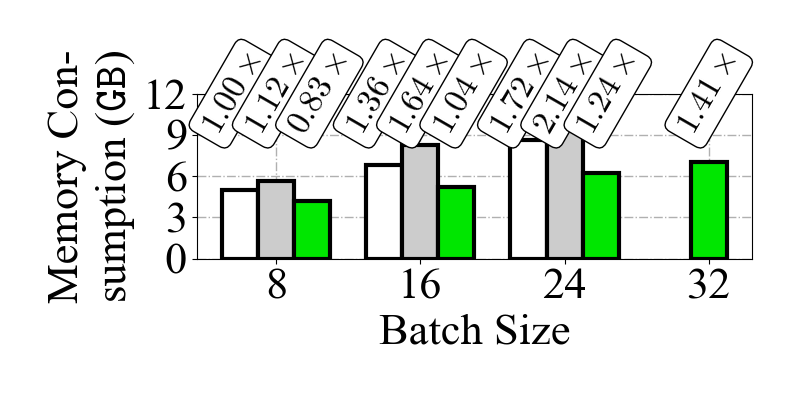}} & 
        \raisebox{-0.5\height}{\includegraphics[width=0.46\linewidth]{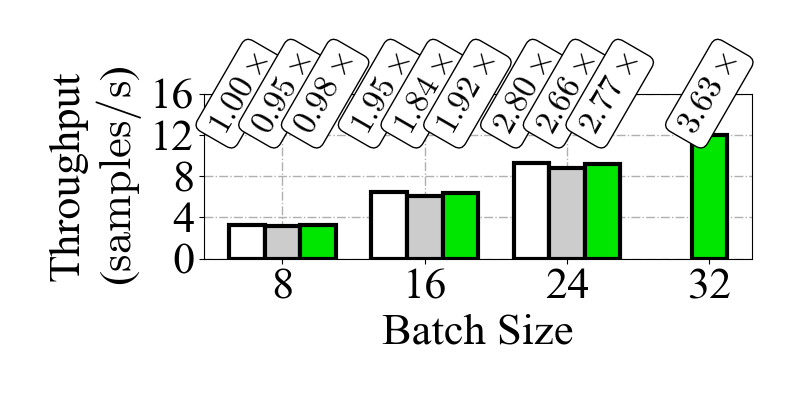}} \\
        \vspace{-5pt} \\
        \raisebox{-0.5\height}{\rotatebox{90}{\footnotesize (2) TX}} &
        \raisebox{-0.5\height}{\includegraphics[width=0.46\linewidth]{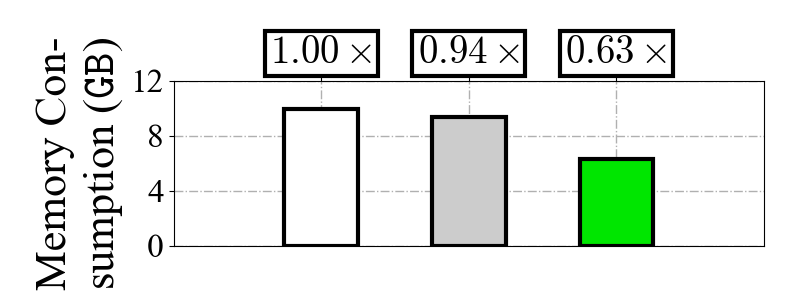}} & 
        \raisebox{-0.5\height}{\includegraphics[width=0.46\linewidth]{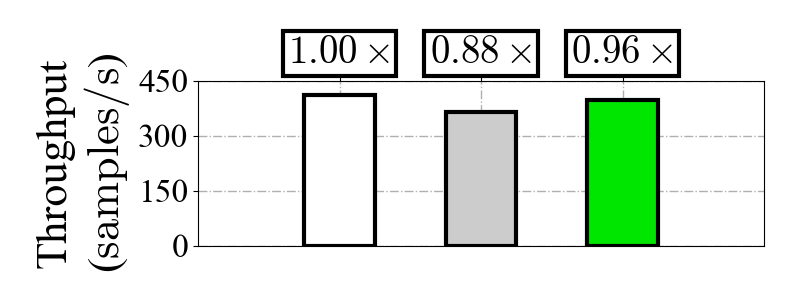}} \\
        \vspace{-5pt} \\
        \raisebox{-0.5\height}{\rotatebox{90}{\footnotesize (3) ResNet}} &
        \raisebox{-0.5\height}{\includegraphics[width=0.46\linewidth]{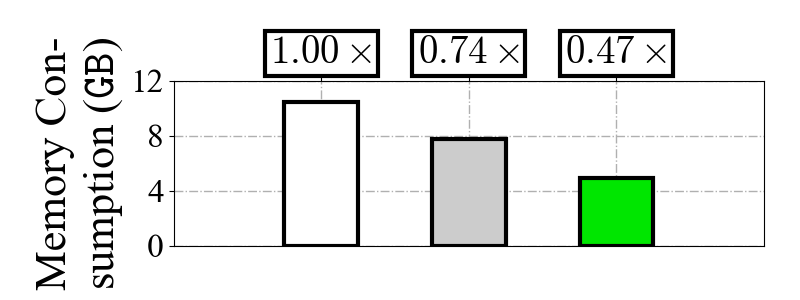}} & 
        \raisebox{-0.5\height}{\includegraphics[width=0.46\linewidth]{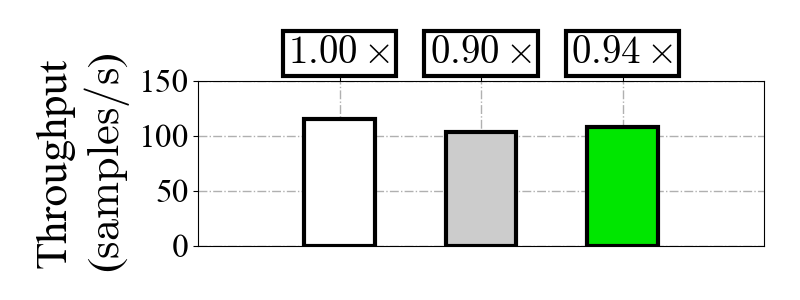}}
    \end{tabular}
    
    \caption{(a) GPU Memory Consumption and (b) Throughput \\
        compared between \Baseline{}, \Mirror{}, and \Echo{} on DS2 (1), 
        Transformer (TX, 2), and ResNet (3)}
    \label{fig:generality}
\end{figure}

Figure~\ref{fig:generality}(1) shows the comparison on the GPU memory consumption and training throughput by training the DeepSpeech2 (DS2) model \cite{deepspeech2}, which is the SOTA model for speech recognition \cite{mlperf, tbd}. We use \(40\mathrm{K}\) audio samples from the LibriSpeech \cite{librispeech} dataset for training.

We observe from Figure~\ref{fig:generality}(1a) that when the batch size is small, \Echo{} cannot provide significant footprint reduction over the baseline. The reason is because in DS2 \cite{deepspeech2} \(2.93\ \mathtt{GB}\) of the GPU memory is allocated for weights, making feature maps less important under small batch size. However, as the batch size increases from \(8\) to \(24\), the GPU memory footprint with \Echo{} increases much slower compared with the baseline and \Mirror{}, because relative proportion of feature maps increases and the latter two cannot adequately address this increase. Thereby, \Echo{} does not hit the GPU memory capacity wall even under a batch size of \(32\), allowing the training throughput to further scale up, as is illustrated in Figure~\ref{fig:generality}(1b). The recomputation runtime overhead of \Echo{} is within \(2\%\) of the baseline under the same batch size, giving it the opportunity to compensate for the performance loss by increasing the batch size. As the rightmost bar of Figure~\ref{fig:generality}(1b) shows, the throughput of \Echo{}\(_{B=32}\) is \(3.6\times\) that of \Baseline{}\(_{B=8}\) and \(1.3\times\) that of \Baseline{}\(_{B=24}\) (the best throughput in the baseline).

We further observe from Figure~\ref{fig:generality}(1a) that \Mirror{} consumes more (rather than less) GPU memory than the baseline. 
This is because it fails to accurately estimate the footprint reduction effects after recomputation (Challenge \#1, Section~\ref{subsec:key_ideas:footprint}).
It also has \(5\times\) more runtime overhead compared to \Echo{}.

\subsubsection{Transformer}\label{subsubsec:evaluation:generality:transformer}
Both the NMT and DS2 are RNN-based models \cite{nmt, gnmt, deepspeech2}. To demonstrate \Echo{}'s generality beyond RNNs, we evaluate the effect of \Echo{} on the Transformer model \cite{transformer}, which is the state-of-the-art model for machine translation \cite{mlperf, tbd} and does not have a RNN component in it. Figure~\ref{fig:generality}(2) shows that \Echo{} achieves a footprint reduction ratio of \TFFootprintReductionRatio{} over the baseline with \(3.0\times\) less overhead than \Mirror{}, where \(38\%\) of the footprint reduction over the baseline comes from layer specific knowledge that the feature maps of the dropout layer can be binarized \cite{gist} (Section~\ref{subsec:key_ideas:runtime}).

\subsubsection{ResNet}\label{subsubsec:evaluation:generality:resnet} 
All the previous models belong to the domain of sequence-to-sequence learning. To show \Echo{}'s generality in other domains, we further evaluate the effect of \Echo{} on the ResNet-152 model, which is the state-of-the-art model for image classification \cite{mlperf, tbd}. Figure~\ref{fig:generality}(3) shows that \Echo{} achieves a footprint ratio of \ResNetFootprintReductionRatio{} over the baseline and \(1.57\times\) over \Mirror{} with \(1.67\times\) less overhead than \Mirror{}.

\CHANGE{Although models such as Transformer and ResNet might not be always able to directly benefit from the footprint reduction by achieving performance gains, they can still benefit indirectly by becoming deeper under the same GPU memory budget. Table~\ref{tab:resnet_and_transformer-depth} shows the maximum number of Transformer and ResNet layers that we can run on one RTX 2080 Ti~\cite{2080-ti}. 
We choose the number of layers specified in \citet{sockeye} and \citet{resnet} and pick the maximum batch sizes that can fit into the \(11\ \mathtt{GB}\) GPU memory budget in the baseline. As we experiment on \Mirror{}, \Echo{} while keeping the batch sizes fixed, we observe that \Echo{} increases the maximum number of layers by \TFMaxNLayers{} and \ResNetMaxNLayers{} on Transformer and ResNet respectively. Such increase in depth
aligns with the recent trends in deep learning that have significant demand for more layers \cite{imagenet, red-ai}.
}

\begin{table}[ht]\CHANGE{
    \centering
    \footnotesize
    \begin{tabular}{lccc}
        \toprule
        \textbf{Model} & \textbf{Baseline} & \textbf{MXNet} & \textbf{Echo} \\
        \midrule
        \textbf{Transformer}
            & \(6\) & \(6\) & \(11\) \\
        \textbf{ResNet}\(\)
            & \(50\) & \(101\) & \(200\) \\
        \bottomrule
    \end{tabular}
    \captionsetup{justification=centering}
    \caption{\CHANGE{Maximum Number of Layers of Transformer~\cite{transformer} and 
        ResNet~\cite{resnet} on one RTX 2080 Ti~\cite{2080-ti}}}
    \label{tab:resnet_and_transformer-depth}
}\end{table}

In summary, by evaluating \Echo{} on diverse workloads, we have shown the generality of the ideas behind \Echo{}. 
We also conclude that \Echo{} is able to efficiently and effectively reduce the memory footprint for RNNs~\cite{nmt, gnmt, deepspeech2}, Transformer~\cite{transformer}, and CNNs (ResNet~\cite{resnet}) with very low overhead. \CHANGE{Such footprint benefits can be converted into performance improvements and/or an increase in the number of layers that can be executed while using the same batch size and GPU memory budget.}


%% file: sections/7-Related_Works.tex
\section{Related Work}

In this work, we present \Echo{}, a compiler-based optimization scheme that automatically and transparently reduces the GPU memory footprint used for training across diverse machine learning models without any changes needed to the training source code. \Echo{} finds and addresses two key challenges of selective recomputation that are missing in prior works \cite{sublinear, memory_effi-bptt, hessian-free}.
Compared with other prior works, \Echo{} (i) focuses on training rather than inference \cite{deep_compression, eie, concise_loads, cnvlutin, diannao, eyeriss, stripes, scnn, minerva, scaledeep}, and (ii) requires no domain-specific knowledge of the computation graph structures and/or manual efforts to hand-write CUDA kernels \cite{gist, tf-memory}.

\textbf{Generic Footprint Reduction Techniques.}
vDNN \cite{vdnn} and CDMA \cite{cdma} fit large neural networks in the GPU memory by moving the data between CPU and GPU using offloading and prefetching. The major weaknesses of those approaches are (1) performance loss (\(15\%\) on average \cite{vdnn}), and (2) intense use of PCIe bus, which can be a bottleneck in distributed training. 

\textbf{CNNs and/or Inference.}
Gist \cite{gist} proposes lossless memory compression technique, but in the context of CNNs. Their technique, however, requires the use of \(\operatorname{ReLU}\) activations. Because LSTM RNNs mostly use \(\operatorname{tanh}\) and \(\operatorname{sigmoid}\), \emph{Gist} encodings become mostly inapplicable in the LSTM RNN context. 
There have been numerous efficient footprint reduction techniques that target inference \cite{deep_compression, eie, concise_loads, cnvlutin, diannao, eyeriss, stripes, scnn, minerva, scaledeep}. Those works mostly focus on reducing the footprint of model weights. However, in training weights are frequently updated, so it is challenging to apply these ideas in the training context. Furthermore, as is shown in Section~\ref{subsec:motivation:memory} (Figure~\ref{fig:sockeye-memory_profile}), weights are not the major memory consumer in NMT training.

\textbf{Algorithm Changes.}
\Echo{} makes no changes to the underlying algorithms of the training models. This makes it different from those approaches that address the memory footprint issue from the algorithmic perspective. For example, \citet{reversible} and \citet{memory_effi-bptt} reduce the memory footprint of RNNs via algorithmic innovations. 

\textbf{Machine Learning Compilers.} \Echo{} is a compiler-based optimization scheme on reducing the memory footprint of DNN training tasks. It is therefore orthogonal to prior works on machine learning compilers that propose new programming paradigm and/or focus on runtime performance improvements, which include TVM~\cite{tvm}, Relay~\cite{relay}, Latte~\cite{latte}, TensorFlow XLA~\cite{xla}, Glow~\cite{glow}, and TensorComprehensions~\cite{tc}.



%% file: sections/8-Conclusions.tex
\section{Conclusion}

In this paper, we propose \Echo{}, a compiler-based optimization scheme that reduces memory footprint automatically and transparently. On four state-of-the-art machine learning models, \CHANGE{\Echo{} reduces the GPU memory footprint by \AVGFootprintReductionRatio{} on average and \NMTFootprintReductionRatio{} maximum with marginal runtime overhead}. System researchers, machine learning practitioners can all benefit from \Echo{}, as it speeds up training convergence, reduces the GPU energy consumption, and allows for larger and deeper models.
We hope that \Echo{} would become an efficient platform for further research on memory footprint optimizations and efficient system design for key machine learning applications.

%% file: main.bbl

\begin{thebibliography}{78}


\ifx \showCODEN    \undefined \def \showCODEN     #1{\unskip}     \fi
\ifx \showDOI      \undefined \def \showDOI       #1{#1}\fi
\ifx \showISBNx    \undefined \def \showISBNx     #1{\unskip}     \fi
\ifx \showISBNxiii \undefined \def \showISBNxiii  #1{\unskip}     \fi
\ifx \showISSN     \undefined \def \showISSN      #1{\unskip}     \fi
\ifx \showLCCN     \undefined \def \showLCCN      #1{\unskip}     \fi
\ifx \shownote     \undefined \def \shownote      #1{#1}          \fi
\ifx \showarticletitle \undefined \def \showarticletitle #1{#1}   \fi
\ifx \showURL      \undefined \def \showURL       {\relax}        \fi
\providecommand\bibfield[2]{#2}
\providecommand\bibinfo[2]{#2}
\providecommand\natexlab[1]{#1}
\providecommand\showeprint[2][]{arXiv:#2}

\bibitem[\protect\citeauthoryear{Albericio, Judd, Hetherington, Aamodt, Jerger,
  and Moshovos}{Albericio et~al\mbox{.}}{2016}]%
        {cnvlutin}
\bibfield{author}{\bibinfo{person}{Jorge Albericio}, \bibinfo{person}{Patrick
  Judd}, \bibinfo{person}{Tayler Hetherington}, \bibinfo{person}{Tor Aamodt},
  \bibinfo{person}{Natalie~Enright Jerger}, {and} \bibinfo{person}{Andreas
  Moshovos}.} \bibinfo{year}{2016}\natexlab{}.
\newblock \showarticletitle{Cnvlutin: Ineffectual-neuron-free Deep Neural
  Network Computing}. In \bibinfo{booktitle}{\emph{Proceedings of the 43rd
  International Symposium on Computer Architecture}}
  \emph{(\bibinfo{series}{ISCA '16})}. \bibinfo{publisher}{IEEE Press},
  \bibinfo{address}{Piscataway, NJ, USA}, \bibinfo{pages}{1--13}.
\newblock
\showISBNx{978-1-4673-8947-1}
\urldef\tempurl%
\url{https://doi.org/10.1109/ISCA.2016.11}
\showDOI{\tempurl}


\bibitem[\protect\citeauthoryear{AMD}{AMD}{2019}]%
        {amd-epyc-7371}
\bibfield{author}{\bibinfo{person}{AMD}.} \bibinfo{year}{2019}\natexlab{}.
\newblock \bibinfo{booktitle}{\emph{AMD EPYC\texttrademark 7371}}.
\newblock
\urldef\tempurl%
\url{https://www.amd.com/en/products/cpu/amd-epyc-7371}
\showURL{%
\tempurl}


\bibitem[\protect\citeauthoryear{Amodei, Anubhai, Battenberg, Case, Casper,
  Catanzaro, Chen, Chrzanowski, Coates, Diamos, Elsen, Engel, Fan, Fougner,
  Han, Hannun, Jun, LeGresley, Lin, Narang, Ng, Ozair, Prenger, Raiman,
  Satheesh, Seetapun, Sengupta, Wang, Wang, Wang, Xiao, Yogatama, Zhan, and
  Zhu}{Amodei et~al\mbox{.}}{2015}]%
        {deepspeech2}
\bibfield{author}{\bibinfo{person}{Dario Amodei}, \bibinfo{person}{Rishita
  Anubhai}, \bibinfo{person}{Eric Battenberg}, \bibinfo{person}{Carl Case},
  \bibinfo{person}{Jared Casper}, \bibinfo{person}{Bryan Catanzaro},
  \bibinfo{person}{Jingdong Chen}, \bibinfo{person}{Mike Chrzanowski},
  \bibinfo{person}{Adam Coates}, \bibinfo{person}{Greg Diamos},
  \bibinfo{person}{Erich Elsen}, \bibinfo{person}{Jesse~H. Engel},
  \bibinfo{person}{Linxi Fan}, \bibinfo{person}{Christopher Fougner},
  \bibinfo{person}{Tony Han}, \bibinfo{person}{Awni~Y. Hannun},
  \bibinfo{person}{Billy Jun}, \bibinfo{person}{Patrick LeGresley},
  \bibinfo{person}{Libby Lin}, \bibinfo{person}{Sharan Narang},
  \bibinfo{person}{Andrew~Y. Ng}, \bibinfo{person}{Sherjil Ozair},
  \bibinfo{person}{Ryan Prenger}, \bibinfo{person}{Jonathan Raiman},
  \bibinfo{person}{Sanjeev Satheesh}, \bibinfo{person}{David Seetapun},
  \bibinfo{person}{Shubho Sengupta}, \bibinfo{person}{Yi Wang},
  \bibinfo{person}{Zhiqian Wang}, \bibinfo{person}{Chong Wang},
  \bibinfo{person}{Bo Xiao}, \bibinfo{person}{Dani Yogatama},
  \bibinfo{person}{Jun Zhan}, {and} \bibinfo{person}{Zhenyao Zhu}.}
  \bibinfo{year}{2015}\natexlab{}.
\newblock \showarticletitle{Deep Speech 2: End-to-End Speech Recognition in
  English and Mandarin}.
\newblock \bibinfo{journal}{\emph{CoRR}}  \bibinfo{volume}{abs/1512.02595}
  (\bibinfo{year}{2015}).
\newblock
\showeprint[arxiv]{1512.02595}
\urldef\tempurl%
\url{http://arxiv.org/abs/1512.02595}
\showURL{%
\tempurl}


\bibitem[\protect\citeauthoryear{(AWS)}{(AWS)}{2019}]%
        {p3_8xlarge}
\bibfield{author}{\bibinfo{person}{Amazon Web~Services (AWS)}.}
  \bibinfo{year}{2019}\natexlab{}.
\newblock \bibinfo{booktitle}{\emph{Amazon EC2 P3 Instance Product Details}}.
\newblock
\urldef\tempurl%
\url{https://aws.amazon.com/ec2/instance-types/p3}
\showURL{%
\tempurl}


\bibitem[\protect\citeauthoryear{Bahdanau, Cho, and Bengio}{Bahdanau
  et~al\mbox{.}}{2014}]%
        {nmt}
\bibfield{author}{\bibinfo{person}{Dzmitry Bahdanau},
  \bibinfo{person}{Kyunghyun Cho}, {and} \bibinfo{person}{Yoshua Bengio}.}
  \bibinfo{year}{2014}\natexlab{}.
\newblock \showarticletitle{Neural Machine Translation by Jointly Learning to
  Align and Translate}.
\newblock \bibinfo{journal}{\emph{CoRR}}  \bibinfo{volume}{abs/1409.0473}
  (\bibinfo{year}{2014}).
\newblock
\showeprint[arxiv]{1409.0473}
\urldef\tempurl%
\url{http://arxiv.org/abs/1409.0473}
\showURL{%
\tempurl}


\bibitem[\protect\citeauthoryear{Bradbury, Merity, Xiong, and Socher}{Bradbury
  et~al\mbox{.}}{2017}]%
        {quasi}
\bibfield{author}{\bibinfo{person}{James Bradbury}, \bibinfo{person}{Stephen
  Merity}, \bibinfo{person}{Caiming Xiong}, {and} \bibinfo{person}{Richard
  Socher}.} \bibinfo{year}{2017}\natexlab{}.
\newblock \showarticletitle{Quasi-Recurrent Neural Networks}.
\newblock \bibinfo{journal}{\emph{International Conference on Learning
  Representations (ICLR 2017)}} (\bibinfo{year}{2017}).
\newblock


\bibitem[\protect\citeauthoryear{Britz, Goldie, Luong, and Le}{Britz
  et~al\mbox{.}}{2017}]%
        {nmt-massive_exploration}
\bibfield{author}{\bibinfo{person}{Denny Britz}, \bibinfo{person}{Anna Goldie},
  \bibinfo{person}{Thang Luong}, {and} \bibinfo{person}{Quoc Le}.}
  \bibinfo{year}{2017}\natexlab{}.
\newblock \showarticletitle{Massive Exploration of Neural Machine Translation
  Architectures}.
\newblock \bibinfo{journal}{\emph{ArXiv e-prints}} (\bibinfo{date}{March}
  \bibinfo{year}{2017}).
\newblock
\showeprint[arxiv]{cs.CL/1703.03906}


\bibitem[\protect\citeauthoryear{Chen, Du, Sun, Wang, Wu, Chen, and Temam}{Chen
  et~al\mbox{.}}{2014}]%
        {diannao}
\bibfield{author}{\bibinfo{person}{Tianshi Chen}, \bibinfo{person}{Zidong Du},
  \bibinfo{person}{Ninghui Sun}, \bibinfo{person}{Jia Wang},
  \bibinfo{person}{Chengyong Wu}, \bibinfo{person}{Yunji Chen}, {and}
  \bibinfo{person}{Olivier Temam}.} \bibinfo{year}{2014}\natexlab{}.
\newblock \showarticletitle{DianNao: A Small-footprint High-throughput
  Accelerator for Ubiquitous Machine-learning}. In
  \bibinfo{booktitle}{\emph{Proceedings of the 19th International Conference on
  Architectural Support for Programming Languages and Operating Systems}}
  \emph{(\bibinfo{series}{ASPLOS '14})}. \bibinfo{publisher}{ACM},
  \bibinfo{address}{New York, NY, USA}, \bibinfo{pages}{269--284}.
\newblock
\showISBNx{978-1-4503-2305-5}
\urldef\tempurl%
\url{https://doi.org/10.1145/2541940.2541967}
\showDOI{\tempurl}


\bibitem[\protect\citeauthoryear{Chen, Li, Li, Lin, Wang, Wang, Xiao, Xu,
  Zhang, and Zhang}{Chen et~al\mbox{.}}{2015}]%
        {mxnet}
\bibfield{author}{\bibinfo{person}{Tianqi Chen}, \bibinfo{person}{Mu Li},
  \bibinfo{person}{Yutian Li}, \bibinfo{person}{Min Lin},
  \bibinfo{person}{Naiyan Wang}, \bibinfo{person}{Minjie Wang},
  \bibinfo{person}{Tianjun Xiao}, \bibinfo{person}{Bing Xu},
  \bibinfo{person}{Chiyuan Zhang}, {and} \bibinfo{person}{Zheng Zhang}.}
  \bibinfo{year}{2015}\natexlab{}.
\newblock \showarticletitle{MXNet: A Flexible and Efficient Machine Learning
  Library for Heterogeneous Distributed Systems}.
\newblock \bibinfo{journal}{\emph{CoRR}}  \bibinfo{volume}{abs/1512.01274}
  (\bibinfo{year}{2015}).
\newblock
\showeprint[arxiv]{1512.01274}
\urldef\tempurl%
\url{http://arxiv.org/abs/1512.01274}
\showURL{%
\tempurl}


\bibitem[\protect\citeauthoryear{Chen, Moreau, Jiang, Zheng, Yan, Shen, Cowan,
  Wang, Hu, Ceze, Guestrin, and Krishnamurthy}{Chen et~al\mbox{.}}{2018}]%
        {tvm}
\bibfield{author}{\bibinfo{person}{Tianqi Chen}, \bibinfo{person}{Thierry
  Moreau}, \bibinfo{person}{Ziheng Jiang}, \bibinfo{person}{Lianmin Zheng},
  \bibinfo{person}{Eddie Yan}, \bibinfo{person}{Haichen Shen},
  \bibinfo{person}{Meghan Cowan}, \bibinfo{person}{Leyuan Wang},
  \bibinfo{person}{Yuwei Hu}, \bibinfo{person}{Luis Ceze},
  \bibinfo{person}{Carlos Guestrin}, {and} \bibinfo{person}{Arvind
  Krishnamurthy}.} \bibinfo{year}{2018}\natexlab{}.
\newblock \showarticletitle{TVM: An Automated End-to-End Optimizing Compiler
  for Deep Learning}. In \bibinfo{booktitle}{\emph{13th USENIX Symposium on
  Operating Systems Design and Implementation (OSDI 18)}}.
  \bibinfo{pages}{578--594}.
\newblock


\bibitem[\protect\citeauthoryear{Chen, Xu, Zhang, and Guestrin}{Chen
  et~al\mbox{.}}{2016b}]%
        {sublinear}
\bibfield{author}{\bibinfo{person}{Tianqi Chen}, \bibinfo{person}{Bing Xu},
  \bibinfo{person}{Chiyuan Zhang}, {and} \bibinfo{person}{Carlos Guestrin}.}
  \bibinfo{year}{2016}\natexlab{b}.
\newblock \showarticletitle{Training Deep Nets with Sublinear Memory Cost}.
\newblock \bibinfo{journal}{\emph{ArXiv e-prints}} (\bibinfo{year}{2016}).
\newblock
\showeprint[arxiv]{cs.LG/1604.06174}


\bibitem[\protect\citeauthoryear{Chen, Emer, and Sze}{Chen
  et~al\mbox{.}}{2016a}]%
        {eyeriss}
\bibfield{author}{\bibinfo{person}{Yu-Hsin Chen}, \bibinfo{person}{Joel Emer},
  {and} \bibinfo{person}{Vivienne Sze}.} \bibinfo{year}{2016}\natexlab{a}.
\newblock \showarticletitle{Eyeriss: A Spatial Architecture for
  Energy-efficient Dataflow for Convolutional Neural Networks}. In
  \bibinfo{booktitle}{\emph{Proceedings of the 43rd International Symposium on
  Computer Architecture}} \emph{(\bibinfo{series}{ISCA '16})}.
  \bibinfo{publisher}{IEEE Press}, \bibinfo{address}{Piscataway, NJ, USA},
  \bibinfo{pages}{367--379}.
\newblock
\showISBNx{978-1-4673-8947-1}
\urldef\tempurl%
\url{https://doi.org/10.1109/ISCA.2016.40}
\showDOI{\tempurl}


\bibitem[\protect\citeauthoryear{CybertronAI}{CybertronAI}{2019}]%
        {openai-grad_ckpt}
\bibfield{author}{\bibinfo{person}{CybertronAI}.}
  \bibinfo{year}{2019}\natexlab{}.
\newblock \bibinfo{booktitle}{\emph{Saving memory using
  gradient-checkpointing}}.
\newblock
\urldef\tempurl%
\url{https://github.com/cybertronai/gradient-checkpointing}
\showURL{%
\tempurl}


\bibitem[\protect\citeauthoryear{Deng, Dong, Socher, Li, Li, and Fei-Fei}{Deng
  et~al\mbox{.}}{2009}]%
        {imagenet}
\bibfield{author}{\bibinfo{person}{Jia Deng}, \bibinfo{person}{Wei Dong},
  \bibinfo{person}{Richard Socher}, \bibinfo{person}{Li-Jia Li},
  \bibinfo{person}{Kai Li}, {and} \bibinfo{person}{Li Fei-Fei}.}
  \bibinfo{year}{2009}\natexlab{}.
\newblock \showarticletitle{Imagenet: A large-scale hierarchical image
  database}. In \bibinfo{booktitle}{\emph{2009 IEEE Conference on Computer
  Vision and Pattern Recognition (CVPR)}}. IEEE, \bibinfo{pages}{248--255}.
\newblock


\bibitem[\protect\citeauthoryear{Graves and Jaitly}{Graves and Jaitly}{2014}]%
        {end2end-speech}
\bibfield{author}{\bibinfo{person}{Alex Graves} {and} \bibinfo{person}{Navdeep
  Jaitly}.} \bibinfo{year}{2014}\natexlab{}.
\newblock \showarticletitle{Towards End-to-End Speech Recognition with
  Recurrent Neural Networks}. In \bibinfo{booktitle}{\emph{International
  Conference on Machine Learning}}. \bibinfo{pages}{1764--1772}.
\newblock


\bibitem[\protect\citeauthoryear{Graves, Mohamed, and Hinton}{Graves
  et~al\mbox{.}}{2013}]%
        {deepspeech}
\bibfield{author}{\bibinfo{person}{Alex Graves}, \bibinfo{person}{Abdelrahman
  Mohamed}, {and} \bibinfo{person}{Geoffrey Hinton}.}
  \bibinfo{year}{2013}\natexlab{}.
\newblock \showarticletitle{Speech Recognition with Deep Recurrent Neural
  Networks}. In \bibinfo{booktitle}{\emph{IEEE International Conference on
  Acoustics, Speech and Signal Processing (ICASSP) 2013}}. IEEE,
  \bibinfo{pages}{6645--6649}.
\newblock


\bibitem[\protect\citeauthoryear{Gruslys, Munos, Danihelka, Lanctot, and
  Graves}{Gruslys et~al\mbox{.}}{2016}]%
        {memory_effi-bptt}
\bibfield{author}{\bibinfo{person}{Audrunas Gruslys},
  \bibinfo{person}{R{\'{e}}mi Munos}, \bibinfo{person}{Ivo Danihelka},
  \bibinfo{person}{Marc Lanctot}, {and} \bibinfo{person}{Alex Graves}.}
  \bibinfo{year}{2016}\natexlab{}.
\newblock \showarticletitle{Memory-Efficient Backpropagation Through Time}.
\newblock \bibinfo{journal}{\emph{CoRR}}  \bibinfo{volume}{abs/1606.03401}
  (\bibinfo{year}{2016}).
\newblock
\showeprint[arxiv]{1606.03401}
\urldef\tempurl%
\url{http://arxiv.org/abs/1606.03401}
\showURL{%
\tempurl}


\bibitem[\protect\citeauthoryear{Han, Liu, Mao, Pu, Pedram, Horowitz, and
  Dally}{Han et~al\mbox{.}}{2016a}]%
        {eie}
\bibfield{author}{\bibinfo{person}{Song Han}, \bibinfo{person}{Xingyu Liu},
  \bibinfo{person}{Huizi Mao}, \bibinfo{person}{Jing Pu},
  \bibinfo{person}{Ardavan Pedram}, \bibinfo{person}{Mark~A. Horowitz}, {and}
  \bibinfo{person}{William~J. Dally}.} \bibinfo{year}{2016}\natexlab{a}.
\newblock \showarticletitle{EIE: Efficient Inference Engine on Compressed Deep
  Neural Network}.
\newblock \bibinfo{journal}{\emph{SIGARCH Computer Architecture News}}
  \bibinfo{volume}{44}, \bibinfo{number}{3} (\bibinfo{date}{June}
  \bibinfo{year}{2016}), \bibinfo{pages}{243--254}.
\newblock
\showISSN{0163-5964}
\urldef\tempurl%
\url{https://doi.org/10.1145/3007787.3001163}
\showDOI{\tempurl}


\bibitem[\protect\citeauthoryear{Han, Mao, and Dally}{Han
  et~al\mbox{.}}{2016b}]%
        {deep_compression}
\bibfield{author}{\bibinfo{person}{Song Han}, \bibinfo{person}{Huizi Mao},
  {and} \bibinfo{person}{William~J. Dally}.} \bibinfo{year}{2016}\natexlab{b}.
\newblock \showarticletitle{Deep Compression: Compressing Deep Neural Network
  with Pruning, Trained Quantization and Huffman Coding}.
\newblock \bibinfo{journal}{\emph{International Conference on Learning
  Representations (ICLR 2016)}} (\bibinfo{year}{2016}).
\newblock


\bibitem[\protect\citeauthoryear{He, Zhang, Ren, and Sun}{He
  et~al\mbox{.}}{2015}]%
        {resnet}
\bibfield{author}{\bibinfo{person}{Kaiming He}, \bibinfo{person}{Xiangyu
  Zhang}, \bibinfo{person}{Shaoqing Ren}, {and} \bibinfo{person}{Jian Sun}.}
  \bibinfo{year}{2015}\natexlab{}.
\newblock \showarticletitle{Deep Residual Learning for Image Recognition}.
\newblock \bibinfo{journal}{\emph{CoRR}}  \bibinfo{volume}{abs/1512.03385}
  (\bibinfo{year}{2015}).
\newblock
\showeprint[arxiv]{1512.03385}
\urldef\tempurl%
\url{http://arxiv.org/abs/1512.03385}
\showURL{%
\tempurl}


\bibitem[\protect\citeauthoryear{Hieber, Domhan, Denkowski, Vilar, Sokolov,
  Clifton, and Post}{Hieber et~al\mbox{.}}{2017}]%
        {sockeye}
\bibfield{author}{\bibinfo{person}{Felix Hieber}, \bibinfo{person}{Tobias
  Domhan}, \bibinfo{person}{Michael Denkowski}, \bibinfo{person}{David Vilar},
  \bibinfo{person}{Artem Sokolov}, \bibinfo{person}{Ann Clifton}, {and}
  \bibinfo{person}{Matt Post}.} \bibinfo{year}{2017}\natexlab{}.
\newblock \showarticletitle{Sockeye: {A} Toolkit for Neural Machine
  Translation}.
\newblock \bibinfo{journal}{\emph{CoRR}}  \bibinfo{volume}{abs/1712.05690}
  (\bibinfo{year}{2017}).
\newblock


\bibitem[\protect\citeauthoryear{Hochreiter and Schmidhuber}{Hochreiter and
  Schmidhuber}{1997}]%
        {lstm}
\bibfield{author}{\bibinfo{person}{Sepp Hochreiter} {and}
  \bibinfo{person}{J\"{u}rgen Schmidhuber}.} \bibinfo{year}{1997}\natexlab{}.
\newblock \showarticletitle{Long Short-Term Memory}.
\newblock \bibinfo{journal}{\emph{Neural Compututation}} \bibinfo{volume}{9},
  \bibinfo{number}{8} (\bibinfo{date}{Nov.} \bibinfo{year}{1997}),
  \bibinfo{pages}{1735--1780}.
\newblock
\showISSN{0899-7667}
\urldef\tempurl%
\url{https://doi.org/10.1162/neco.1997.9.8.1735}
\showDOI{\tempurl}


\bibitem[\protect\citeauthoryear{Jain, Hill, Lin, Khan, Haque, Laurenzano,
  Mahlke, Tang, and Mars}{Jain et~al\mbox{.}}{2016}]%
        {concise_loads}
\bibfield{author}{\bibinfo{person}{Animesh Jain}, \bibinfo{person}{Parker
  Hill}, \bibinfo{person}{Shih-Chieh Lin}, \bibinfo{person}{Muneeb Khan},
  \bibinfo{person}{Md~E. Haque}, \bibinfo{person}{Michael~A. Laurenzano},
  \bibinfo{person}{Scott Mahlke}, \bibinfo{person}{Lingjia Tang}, {and}
  \bibinfo{person}{Jason Mars}.} \bibinfo{year}{2016}\natexlab{}.
\newblock \showarticletitle{Concise Loads and Stores: The Case for an
  Asymmetric Compute-memory Architecture for Approximation}. In
  \bibinfo{booktitle}{\emph{The 49th Annual IEEE/ACM International Symposium on
  Microarchitecture}} \emph{(\bibinfo{series}{MICRO-49})}.
  \bibinfo{publisher}{IEEE Press}, \bibinfo{address}{Piscataway, NJ, USA},
  Article \bibinfo{articleno}{41}, \bibinfo{numpages}{13}~pages.
\newblock
\urldef\tempurl%
\url{http://dl.acm.org/citation.cfm?id=3195638.3195688}
\showURL{%
\tempurl}


\bibitem[\protect\citeauthoryear{Jain, Phanishayee, Mars, Tang, and
  Pekhimenko}{Jain et~al\mbox{.}}{2018}]%
        {gist}
\bibfield{author}{\bibinfo{person}{Animesh Jain}, \bibinfo{person}{Amar
  Phanishayee}, \bibinfo{person}{Jason Mars}, \bibinfo{person}{Lingjia Tang},
  {and} \bibinfo{person}{Gennady Pekhimenko}.} \bibinfo{year}{2018}\natexlab{}.
\newblock \showarticletitle{Gist: Efficient Data Encoding for Deep Neural
  Network Training}. In \bibinfo{booktitle}{\emph{Proceeding of the 45st Annual
  International Symposium on Computer Architecture}}
  \emph{(\bibinfo{series}{ISCA 2018})}. \bibinfo{pages}{776--789}.
\newblock
\urldef\tempurl%
\url{https://doi.org/10.1109/ISCA.2018.00070}
\showDOI{\tempurl}


\bibitem[\protect\citeauthoryear{JEDEC}{JEDEC}{2016}]%
        {gddr6}
\bibfield{author}{\bibinfo{person}{JEDEC}.} \bibinfo{year}{2016}\natexlab{}.
\newblock \bibinfo{booktitle}{\emph{Graphics Double Data Rate (GDDR5X) SGRAM
  Standard}}.
\newblock
\urldef\tempurl%
\url{https://www.jedec.org/standards-documents/docs/jesd232a}
\showURL{%
\tempurl}


\bibitem[\protect\citeauthoryear{JEDEC}{JEDEC}{2017}]%
        {ddr4}
\bibfield{author}{\bibinfo{person}{JEDEC}.} \bibinfo{year}{2017}\natexlab{}.
\newblock \bibinfo{booktitle}{\emph{Main Memory: DDR4 and DDR5 SDRAM}}.
\newblock
\urldef\tempurl%
\url{https://www.jedec.org/category/technology-focus-area/main-memory-ddr3-ddr4-sdram}
\showURL{%
\tempurl}


\bibitem[\protect\citeauthoryear{Judd, Albericio, Hetherington, Aamodt, and
  Moshovos}{Judd et~al\mbox{.}}{2016}]%
        {stripes}
\bibfield{author}{\bibinfo{person}{Patrick Judd}, \bibinfo{person}{Jorge
  Albericio}, \bibinfo{person}{Tayler Hetherington}, \bibinfo{person}{Tor~M.
  Aamodt}, {and} \bibinfo{person}{Andreas Moshovos}.}
  \bibinfo{year}{2016}\natexlab{}.
\newblock \showarticletitle{Stripes: Bit-serial Deep Neural Network Computing}.
  In \bibinfo{booktitle}{\emph{The 49th Annual IEEE/ACM International Symposium
  on Microarchitecture}} \emph{(\bibinfo{series}{MICRO-49})}.
  \bibinfo{publisher}{IEEE Press}, \bibinfo{address}{Piscataway, NJ, USA},
  Article \bibinfo{articleno}{19}, \bibinfo{numpages}{12}~pages.
\newblock
\urldef\tempurl%
\url{http://dl.acm.org/citation.cfm?id=3195638.3195661}
\showURL{%
\tempurl}


\bibitem[\protect\citeauthoryear{Klein, Kim, Deng, Senellart, and Rush}{Klein
  et~al\mbox{.}}{2017}]%
        {opennmt}
\bibfield{author}{\bibinfo{person}{Guillaume Klein}, \bibinfo{person}{Yoon
  Kim}, \bibinfo{person}{Yuntian Deng}, \bibinfo{person}{Jean Senellart}, {and}
  \bibinfo{person}{Alexander~M. Rush}.} \bibinfo{year}{2017}\natexlab{}.
\newblock \showarticletitle{OpenNMT: Open-Source Toolkit for Neural Machine
  Translation}.
\newblock \bibinfo{journal}{\emph{CoRR}}  \bibinfo{volume}{abs/1701.02810}
  (\bibinfo{year}{2017}).
\newblock


\bibitem[\protect\citeauthoryear{Labs}{Labs}{2019a}]%
        {cluster-2080_ti}
\bibfield{author}{\bibinfo{person}{Lambda Labs}.}
  \bibinfo{year}{2019}\natexlab{a}.
\newblock \bibinfo{booktitle}{\emph{Deep Learning Workstation for 2019 - 4x
  GPUs: Customize | Lambda Quad}}.
\newblock
\urldef\tempurl%
\url{https://lambdalabs.com/deep-learning/workstations/4-gpu/customize}
\showURL{%
\tempurl}


\bibitem[\protect\citeauthoryear{Labs}{Labs}{2019b}]%
        {cluster-v100}
\bibfield{author}{\bibinfo{person}{Lambda Labs}.}
  \bibinfo{year}{2019}\natexlab{b}.
\newblock \bibinfo{booktitle}{\emph{Tesla V100 Server - 4 GPUs or 8 GPUs +
  NVLink | Lambda Hyperplane}}.
\newblock
\urldef\tempurl%
\url{https://lambdalabs.com/deep-learning/servers/hyperplane/customize}
\showURL{%
\tempurl}


\bibitem[\protect\citeauthoryear{Lei, Zhang, and Artzi}{Lei
  et~al\mbox{.}}{2017}]%
        {sru}
\bibfield{author}{\bibinfo{person}{Tao Lei}, \bibinfo{person}{Yu Zhang}, {and}
  \bibinfo{person}{Yoav Artzi}.} \bibinfo{year}{2017}\natexlab{}.
\newblock \showarticletitle{Training RNNs as Fast as CNNs}.
\newblock \bibinfo{journal}{\emph{CoRR}}  \bibinfo{volume}{abs/1709.02755}
  (\bibinfo{year}{2017}).
\newblock
\showeprint[arxiv]{1709.02755}
\urldef\tempurl%
\url{http://arxiv.org/abs/1709.02755}
\showURL{%
\tempurl}


\bibitem[\protect\citeauthoryear{Luong, Brevdo, and Zhao}{Luong
  et~al\mbox{.}}{2017a}]%
        {tf-nmt}
\bibfield{author}{\bibinfo{person}{Minh{-}Thang Luong}, \bibinfo{person}{Eugene
  Brevdo}, {and} \bibinfo{person}{Rui Zhao}.} \bibinfo{year}{2017}\natexlab{a}.
\newblock \showarticletitle{Neural Machine Translation (seq2seq) Tutorial}.
\newblock  (\bibinfo{year}{2017}).
\newblock
\urldef\tempurl%
\url{https://github.com/tensorflow/nmt}
\showURL{%
\tempurl}


\bibitem[\protect\citeauthoryear{Luong, Brevdo, and Zhao}{Luong
  et~al\mbox{.}}{2017b}]%
        {tf-nmt-multi_gpu}
\bibfield{author}{\bibinfo{person}{Minh{-}Thang Luong}, \bibinfo{person}{Eugene
  Brevdo}, {and} \bibinfo{person}{Rui Zhao}.} \bibinfo{year}{2017}\natexlab{b}.
\newblock \showarticletitle{Neural Machine Translation (seq2seq) Tutorial
  \#Multi-GPU training}.
\newblock  (\bibinfo{year}{2017}).
\newblock
\urldef\tempurl%
\url{https://github.com/tensorflow/nmt}
\showURL{%
\tempurl}


\bibitem[\protect\citeauthoryear{Luong, Brevdo, and Zhao}{Luong
  et~al\mbox{.}}{2017c}]%
        {tf-nmt-wmt16_deen}
\bibfield{author}{\bibinfo{person}{Minh{-}Thang Luong}, \bibinfo{person}{Eugene
  Brevdo}, {and} \bibinfo{person}{Rui Zhao}.} \bibinfo{year}{2017}\natexlab{c}.
\newblock \showarticletitle{Neural Machine Translation (seq2seq) Tutorial \#WMT
  German-English}.
\newblock  (\bibinfo{year}{2017}).
\newblock
\urldef\tempurl%
\url{https://github.com/tensorflow/nmt}
\showURL{%
\tempurl}


\bibitem[\protect\citeauthoryear{MacKay, Vicol, Ba, and Grosse}{MacKay
  et~al\mbox{.}}{2018}]%
        {reversible}
\bibfield{author}{\bibinfo{person}{Matthew MacKay}, \bibinfo{person}{Paul
  Vicol}, \bibinfo{person}{Jimmy Ba}, {and} \bibinfo{person}{Roger~B Grosse}.}
  \bibinfo{year}{2018}\natexlab{}.
\newblock \showarticletitle{Reversible Recurrent Neural Networks}.
\newblock In \bibinfo{booktitle}{\emph{Advances in Neural Information
  Processing Systems 31}}, \bibfield{editor}{\bibinfo{person}{S.~Bengio},
  \bibinfo{person}{H.~Wallach}, \bibinfo{person}{H.~Larochelle},
  \bibinfo{person}{K.~Grauman}, \bibinfo{person}{N.~Cesa-Bianchi}, {and}
  \bibinfo{person}{R.~Garnett}} (Eds.). \bibinfo{publisher}{Curran Associates,
  Inc.}, \bibinfo{pages}{9042--9053}.
\newblock
\urldef\tempurl%
\url{http://papers.nips.cc/paper/8117-reversible-recurrent-neural-networks.pdf}
\showURL{%
\tempurl}


\bibitem[\protect\citeauthoryear{Manning, Luong, See, and Pham}{Manning
  et~al\mbox{.}}{2018}]%
        {iwslt15}
\bibfield{author}{\bibinfo{person}{Christopher~D. Manning},
  \bibinfo{person}{Minh-Thang Luong}, \bibinfo{person}{Abigail See}, {and}
  \bibinfo{person}{Hieu Pham}.} \bibinfo{year}{2018}\natexlab{}.
\newblock \bibinfo{booktitle}{\emph{Neural Machine Translation}}.
\newblock
\urldef\tempurl%
\url{https://nlp.stanford.edu/projects/nmt/}
\showURL{%
\tempurl}


\bibitem[\protect\citeauthoryear{Martens and Sutskever}{Martens and
  Sutskever}{2012}]%
        {hessian-free}
\bibfield{author}{\bibinfo{person}{James Martens} {and} \bibinfo{person}{Ilya
  Sutskever}.} \bibinfo{year}{2012}\natexlab{}.
\newblock \bibinfo{booktitle}{\emph{Training Deep and Recurrent Networks with
  Hessian-Free Optimization}}.
\newblock \bibinfo{publisher}{Springer Berlin Heidelberg},
  \bibinfo{address}{Berlin, Heidelberg}, \bibinfo{pages}{479--535}.
\newblock
\showISBNx{978-3-642-35289-8}
\urldef\tempurl%
\url{https://doi.org/10.1007/978-3-642-35289-8_27}
\showDOI{\tempurl}


\bibitem[\protect\citeauthoryear{Meng, Sun, Yang, Qiu, and Gu}{Meng
  et~al\mbox{.}}{2017}]%
        {tf-memory}
\bibfield{author}{\bibinfo{person}{Chen Meng}, \bibinfo{person}{Minmin Sun},
  \bibinfo{person}{Jun Yang}, \bibinfo{person}{Minghui Qiu}, {and}
  \bibinfo{person}{Yang Gu}.} \bibinfo{year}{2017}\natexlab{}.
\newblock \showarticletitle{Training Deeper Models by GPU Memory Optimization
  on TensorFlow}. In \bibinfo{booktitle}{\emph{Proceedings of Machine Learning
  Systems Workshop in NIPS}}.
\newblock


\bibitem[\protect\citeauthoryear{MLPerf}{MLPerf}{2019}]%
        {mlperf}
\bibfield{author}{\bibinfo{person}{MLPerf}.} \bibinfo{year}{2019}\natexlab{}.
\newblock \bibinfo{booktitle}{\emph{MLPerf Reference Implementations}}.
\newblock
\urldef\tempurl%
\url{https://github.com/mlperf/training}
\showURL{%
\tempurl}


\bibitem[\protect\citeauthoryear{MXNet}{MXNet}{2017}]%
        {mxnet-0121}
\bibfield{author}{\bibinfo{person}{Apache MXNet}.}
  \bibinfo{year}{2017}\natexlab{}.
\newblock \bibinfo{booktitle}{\emph{MXNet ver. 0.12.1}}.
\newblock
\urldef\tempurl%
\url{https://github.com/apache/incubator-mxnet/tree/0.12.1}
\showURL{%
\tempurl}


\bibitem[\protect\citeauthoryear{MXNet}{MXNet}{2018}]%
        {mxnet-speedometer}
\bibfield{author}{\bibinfo{person}{Apache MXNet}.}
  \bibinfo{year}{2018}\natexlab{}.
\newblock \bibinfo{booktitle}{\emph{Speedometer}}.
\newblock
\urldef\tempurl%
\url{https://github.com/apache/incubator-mxnet/blob/master/python/mxnet/callback.py}
\showURL{%
\tempurl}


\bibitem[\protect\citeauthoryear{NNVM}{NNVM}{2017}]%
        {nnvm}
\bibfield{author}{\bibinfo{person}{DMLC NNVM}.}
  \bibinfo{year}{2017}\natexlab{}.
\newblock \bibinfo{booktitle}{\emph{NNVM: Build deep learning system by
  parts}}.
\newblock
\urldef\tempurl%
\url{https://github.com/dmlc/nnvm}
\showURL{%
\tempurl}


\bibitem[\protect\citeauthoryear{NVIDIA}{NVIDIA}{2017}]%
        {v100}
\bibfield{author}{\bibinfo{person}{NVIDIA}.} \bibinfo{year}{2017}\natexlab{}.
\newblock \bibinfo{booktitle}{\emph{Tesla V100 Data Center GPU}}.
\newblock
\urldef\tempurl%
\url{https://www.nvidia.com/en-us/data-center/tesla-v100/}
\showURL{%
\tempurl}


\bibitem[\protect\citeauthoryear{NVIDIA}{NVIDIA}{2018a}]%
        {cuda-10}
\bibfield{author}{\bibinfo{person}{NVIDIA}.} \bibinfo{year}{2018}\natexlab{a}.
\newblock \bibinfo{booktitle}{\emph{CUDA Toolkit Documentation v10.0}}.
\newblock
\urldef\tempurl%
\url{https://docs.nvidia.com/cuda/archive/10.0/}
\showURL{%
\tempurl}


\bibitem[\protect\citeauthoryear{NVIDIA}{NVIDIA}{2018b}]%
        {2080-ti}
\bibfield{author}{\bibinfo{person}{NVIDIA}.} \bibinfo{year}{2018}\natexlab{b}.
\newblock \bibinfo{booktitle}{\emph{GEFORCE RTX 2080 Ti}}.
\newblock
\urldef\tempurl%
\url{https://www.nvidia.com/en-us/geforce/graphics-cards/rtx-2080-ti/}
\showURL{%
\tempurl}


\bibitem[\protect\citeauthoryear{NVIDIA}{NVIDIA}{2018c}]%
        {nvidia-smi}
\bibfield{author}{\bibinfo{person}{NVIDIA}.} \bibinfo{year}{2018}\natexlab{c}.
\newblock \bibinfo{booktitle}{\emph{NVIDIA System Management Interface
  program}}.
\newblock
\urldef\tempurl%
\url{http://developer.download.nvidia.com/compute/DCGM/docs/nvidia-smi-367.38.pdf}
\showURL{%
\tempurl}


\bibitem[\protect\citeauthoryear{NVIDIA}{NVIDIA}{2018d}]%
        {turing}
\bibfield{author}{\bibinfo{person}{NVIDIA}.} \bibinfo{year}{2018}\natexlab{d}.
\newblock \bibinfo{booktitle}{\emph{NVIDIA Turing GPU Architecture
  Whitepaper}}.
\newblock
\urldef\tempurl%
\url{https://www.nvidia.com/content/dam/en-zz/Solutions/design-visualization/technologies/turing-architecture/NVIDIA-Turing-Architecture-Whitepaper.pdf}
\showURL{%
\tempurl}


\bibitem[\protect\citeauthoryear{NVIDIA}{NVIDIA}{2018e}]%
        {nvprof}
\bibfield{author}{\bibinfo{person}{NVIDIA}.} \bibinfo{year}{2018}\natexlab{e}.
\newblock \bibinfo{booktitle}{\emph{Profiler User's Guide v10.0}}.
\newblock
\urldef\tempurl%
\url{https://docs.nvidia.com/cuda/archive/10.0/profiler-users-guide/index.html}
\showURL{%
\tempurl}


\bibitem[\protect\citeauthoryear{NVIDIA}{NVIDIA}{2019a}]%
        {cudnn-7_6_3}
\bibfield{author}{\bibinfo{person}{NVIDIA}.} \bibinfo{year}{2019}\natexlab{a}.
\newblock \bibinfo{booktitle}{\emph{cuDNN Library Developer Guide v7.6.3}}.
\newblock
\urldef\tempurl%
\url{https://docs.nvidia.com/deeplearning/sdk/cudnn-archived/cudnn_763/cudnn-developer-guide/index.html}
\showURL{%
\tempurl}


\bibitem[\protect\citeauthoryear{NVIDIA}{NVIDIA}{2019b}]%
        {nvlink}
\bibfield{author}{\bibinfo{person}{NVIDIA}.} \bibinfo{year}{2019}\natexlab{b}.
\newblock \bibinfo{booktitle}{\emph{NVLink}}.
\newblock
\urldef\tempurl%
\url{https://www.nvidia.com/en-us/data-center/nvlink/}
\showURL{%
\tempurl}


\bibitem[\protect\citeauthoryear{Panayotov, Chen, Povey, and
  Khudanpur}{Panayotov et~al\mbox{.}}{2015}]%
        {librispeech}
\bibfield{author}{\bibinfo{person}{Vassil Panayotov}, \bibinfo{person}{Guoguo
  Chen}, \bibinfo{person}{Daniel Povey}, {and} \bibinfo{person}{Sanjeev
  Khudanpur}.} \bibinfo{year}{2015}\natexlab{}.
\newblock \showarticletitle{Librispeech: an ASR corpus based on public domain
  audio books}. In \bibinfo{booktitle}{\emph{2015 IEEE International Conference
  on Acoustics, Speech and Signal Processing (ICASSP)}}. IEEE,
  \bibinfo{pages}{5206--5210}.
\newblock


\bibitem[\protect\citeauthoryear{Papineni, Roukos, Ward, and Zhu}{Papineni
  et~al\mbox{.}}{2002}]%
        {bleu}
\bibfield{author}{\bibinfo{person}{Kishore Papineni}, \bibinfo{person}{Salim
  Roukos}, \bibinfo{person}{Todd Ward}, {and} \bibinfo{person}{Wei-Jing Zhu}.}
  \bibinfo{year}{2002}\natexlab{}.
\newblock \showarticletitle{BLEU: A Method for Automatic Evaluation of Machine
  Translation}. In \bibinfo{booktitle}{\emph{Proceedings of the 40th Annual
  Meeting on Association for Computational Linguistics}}
  \emph{(\bibinfo{series}{ACL '02})}. \bibinfo{publisher}{Association for
  Computational Linguistics}, \bibinfo{address}{Stroudsburg, PA, USA},
  \bibinfo{pages}{311--318}.
\newblock
\urldef\tempurl%
\url{https://doi.org/10.3115/1073083.1073135}
\showDOI{\tempurl}


\bibitem[\protect\citeauthoryear{Parashar, Rhu, Mukkara, Puglielli, Venkatesan,
  Khailany, Emer, Keckler, and Dally}{Parashar et~al\mbox{.}}{2017}]%
        {scnn}
\bibfield{author}{\bibinfo{person}{Angshuman Parashar}, \bibinfo{person}{Minsoo
  Rhu}, \bibinfo{person}{Anurag Mukkara}, \bibinfo{person}{Antonio Puglielli},
  \bibinfo{person}{Rangharajan Venkatesan}, \bibinfo{person}{Brucek Khailany},
  \bibinfo{person}{Joel Emer}, \bibinfo{person}{Stephen~W. Keckler}, {and}
  \bibinfo{person}{William~J. Dally}.} \bibinfo{year}{2017}\natexlab{}.
\newblock \showarticletitle{SCNN: An Accelerator for Compressed-sparse
  Convolutional Neural Networks}. In \bibinfo{booktitle}{\emph{Proceedings of
  the 44th Annual International Symposium on Computer Architecture}}
  \emph{(\bibinfo{series}{ISCA 2017})}. \bibinfo{publisher}{ACM},
  \bibinfo{address}{New York, NY, USA}, \bibinfo{pages}{27--40}.
\newblock
\showISBNx{978-1-4503-4892-8}
\urldef\tempurl%
\url{https://doi.org/10.1145/3079856.3080254}
\showDOI{\tempurl}


\bibitem[\protect\citeauthoryear{Reagen, Whatmough, Adolf, Rama, Lee, Lee,
  Hern\'{a}ndez-Lobato, Wei, and Brooks}{Reagen et~al\mbox{.}}{2016}]%
        {minerva}
\bibfield{author}{\bibinfo{person}{Brandon Reagen}, \bibinfo{person}{Paul
  Whatmough}, \bibinfo{person}{Robert Adolf}, \bibinfo{person}{Saketh Rama},
  \bibinfo{person}{Hyunkwang Lee}, \bibinfo{person}{Sae~Kyu Lee},
  \bibinfo{person}{Jos{\'e}~Miguel Hern\'{a}ndez-Lobato},
  \bibinfo{person}{Gu-Yeon Wei}, {and} \bibinfo{person}{David Brooks}.}
  \bibinfo{year}{2016}\natexlab{}.
\newblock \showarticletitle{Minerva: Enabling Low-power, Highly-accurate Deep
  Neural Network Accelerators}. In \bibinfo{booktitle}{\emph{Proceedings of the
  43rd International Symposium on Computer Architecture}}
  \emph{(\bibinfo{series}{ISCA '16})}. \bibinfo{publisher}{IEEE Press},
  \bibinfo{address}{Piscataway, NJ, USA}, \bibinfo{pages}{267--278}.
\newblock
\showISBNx{978-1-4673-8947-1}
\urldef\tempurl%
\url{https://doi.org/10.1109/ISCA.2016.32}
\showDOI{\tempurl}


\bibitem[\protect\citeauthoryear{Research}{Research}{2018}]%
        {google-bert}
\bibfield{author}{\bibinfo{person}{Google~AI Research}.}
  \bibinfo{year}{2018}\natexlab{}.
\newblock \bibinfo{booktitle}{\emph{BERT \#Out-of-memory issues}}.
\newblock
\urldef\tempurl%
\url{https://github.com/google-research/bert#out-of-memory-issues}
\showURL{%
\tempurl}


\bibitem[\protect\citeauthoryear{Rhu, Gimelshein, Clemons, Zulfiqar, and
  Keckler}{Rhu et~al\mbox{.}}{2016}]%
        {vdnn}
\bibfield{author}{\bibinfo{person}{Minsoo Rhu}, \bibinfo{person}{Natalia
  Gimelshein}, \bibinfo{person}{Jason Clemons}, \bibinfo{person}{Arslan
  Zulfiqar}, {and} \bibinfo{person}{Stephen~W. Keckler}.}
  \bibinfo{year}{2016}\natexlab{}.
\newblock \showarticletitle{vDNN: Virtualized Deep Neural Networks for
  Scalable, Memory-efficient Neural Network Design}. In
  \bibinfo{booktitle}{\emph{The 49th Annual IEEE/ACM International Symposium on
  Microarchitecture}} \emph{(\bibinfo{series}{MICRO-49})}.
  \bibinfo{publisher}{IEEE Press}, \bibinfo{address}{Piscataway, NJ, USA},
  Article \bibinfo{articleno}{18}, \bibinfo{numpages}{13}~pages.
\newblock
\urldef\tempurl%
\url{http://dl.acm.org/citation.cfm?id=3195638.3195660}
\showURL{%
\tempurl}


\bibitem[\protect\citeauthoryear{Rhu, O'Connor, Chatterjee, Pool, Kwon, and
  Keckler}{Rhu et~al\mbox{.}}{2018}]%
        {cdma}
\bibfield{author}{\bibinfo{person}{Minsoo Rhu}, \bibinfo{person}{Mike
  O'Connor}, \bibinfo{person}{Niladrish Chatterjee}, \bibinfo{person}{Jeff
  Pool}, \bibinfo{person}{Youngeun Kwon}, {and} \bibinfo{person}{Stephen~W.
  Keckler}.} \bibinfo{year}{2018}\natexlab{}.
\newblock \showarticletitle{Compressing {DMA} Engine: Leveraging Activation
  Sparsity for Training Deep Neural Networks}. In
  \bibinfo{booktitle}{\emph{{IEEE} International Symposium on High Performance
  Computer Architecture, {HPCA} 2018, Vienna, Austria, February 24-28, 2018}}.
  \bibinfo{publisher}{{IEEE} Computer Society}, \bibinfo{pages}{78--91}.
\newblock
\showISBNx{978-1-5386-3659-6}
\urldef\tempurl%
\url{https://doi.org/10.1109/HPCA.2018.00017}
\showDOI{\tempurl}


\bibitem[\protect\citeauthoryear{Roesch, Lyubomirsky, Weber, Pollock, Kirisame,
  Chen, and Tatlock}{Roesch et~al\mbox{.}}{2018}]%
        {relay}
\bibfield{author}{\bibinfo{person}{Jared Roesch}, \bibinfo{person}{Steven
  Lyubomirsky}, \bibinfo{person}{Logan Weber}, \bibinfo{person}{Josh Pollock},
  \bibinfo{person}{Marisa Kirisame}, \bibinfo{person}{Tianqi Chen}, {and}
  \bibinfo{person}{Zachary Tatlock}.} \bibinfo{year}{2018}\natexlab{}.
\newblock \showarticletitle{Relay: A New IR for Machine Learning Frameworks}.
  In \bibinfo{booktitle}{\emph{Proceedings of the 2Nd ACM SIGPLAN International
  Workshop on Machine Learning and Programming Languages}}
  \emph{(\bibinfo{series}{MAPL 2018})}. \bibinfo{publisher}{ACM},
  \bibinfo{address}{New York, NY, USA}, \bibinfo{pages}{58--68}.
\newblock
\showISBNx{978-1-4503-5834-7}
\urldef\tempurl%
\url{https://doi.org/10.1145/3211346.3211348}
\showDOI{\tempurl}


\bibitem[\protect\citeauthoryear{Rotem, Fix, Abdulrasool, Deng, Dzhabarov,
  Hegeman, Levenstein, Maher, Satish, Olesen, Park, Rakhov, and
  Smelyanskiy}{Rotem et~al\mbox{.}}{2018}]%
        {glow}
\bibfield{author}{\bibinfo{person}{Nadav Rotem}, \bibinfo{person}{Jordan Fix},
  \bibinfo{person}{Saleem Abdulrasool}, \bibinfo{person}{Summer Deng},
  \bibinfo{person}{Roman Dzhabarov}, \bibinfo{person}{James Hegeman},
  \bibinfo{person}{Roman Levenstein}, \bibinfo{person}{Bert Maher},
  \bibinfo{person}{Nadathur Satish}, \bibinfo{person}{Jakob Olesen},
  \bibinfo{person}{Jongsoo Park}, \bibinfo{person}{Artem Rakhov}, {and}
  \bibinfo{person}{Misha Smelyanskiy}.} \bibinfo{year}{2018}\natexlab{}.
\newblock \showarticletitle{Glow: Graph Lowering Compiler Techniques for Neural
  Networks}.
\newblock \bibinfo{journal}{\emph{CoRR}}  \bibinfo{volume}{abs/1805.00907}
  (\bibinfo{year}{2018}).
\newblock
\showeprint[arxiv]{1805.00907}
\urldef\tempurl%
\url{http://arxiv.org/abs/1805.00907}
\showURL{%
\tempurl}


\bibitem[\protect\citeauthoryear{Rumelhart, Hinton, and Williams}{Rumelhart
  et~al\mbox{.}}{1986}]%
        {backprop}
\bibfield{author}{\bibinfo{person}{D.~E. Rumelhart}, \bibinfo{person}{G.~E.
  Hinton}, {and} \bibinfo{person}{R.~J. Williams}.}
  \bibinfo{year}{1986}\natexlab{}.
\newblock \showarticletitle{Parallel Distributed Processing: Explorations in
  the Microstructure of Cognition, Vol. 1}.
\newblock \bibinfo{publisher}{MIT Press}, \bibinfo{address}{Cambridge, MA,
  USA}, Chapter Learning Internal Representations by Error Propagation,
  \bibinfo{pages}{318--362}.
\newblock
\showISBNx{0-262-68053-X}
\urldef\tempurl%
\url{http://dl.acm.org/citation.cfm?id=104279.104293}
\showURL{%
\tempurl}


\bibitem[\protect\citeauthoryear{Schwartz, Dodge, Smith, and Etzioni}{Schwartz
  et~al\mbox{.}}{2019}]%
        {red-ai}
\bibfield{author}{\bibinfo{person}{Roy Schwartz}, \bibinfo{person}{Jesse
  Dodge}, \bibinfo{person}{Noah~A. Smith}, {and} \bibinfo{person}{Oren
  Etzioni}.} \bibinfo{year}{2019}\natexlab{}.
\newblock \showarticletitle{Green {AI}}.
\newblock \bibinfo{journal}{\emph{CoRR}}  \bibinfo{volume}{abs/1907.10597}
  (\bibinfo{year}{2019}).
\newblock
\showeprint[arxiv]{1907.10597}
\urldef\tempurl%
\url{http://arxiv.org/abs/1907.10597}
\showURL{%
\tempurl}


\bibitem[\protect\citeauthoryear{Srivastava, Hinton, Krizhevsky, Sutskever, and
  Salakhutdinov}{Srivastava et~al\mbox{.}}{2014}]%
        {dropout}
\bibfield{author}{\bibinfo{person}{Nitish Srivastava},
  \bibinfo{person}{Geoffrey Hinton}, \bibinfo{person}{Alex Krizhevsky},
  \bibinfo{person}{Ilya Sutskever}, {and} \bibinfo{person}{Ruslan
  Salakhutdinov}.} \bibinfo{year}{2014}\natexlab{}.
\newblock \showarticletitle{Dropout: A Simple Way to Prevent Neural Networks
  from Overfitting}.
\newblock \bibinfo{journal}{\emph{J. Mach. Learn. Res.}} \bibinfo{volume}{15},
  \bibinfo{number}{1} (\bibinfo{date}{Jan.} \bibinfo{year}{2014}),
  \bibinfo{pages}{1929--1958}.
\newblock
\showISSN{1532-4435}
\urldef\tempurl%
\url{http://dl.acm.org/citation.cfm?id=2627435.2670313}
\showURL{%
\tempurl}


\bibitem[\protect\citeauthoryear{Sundermeyer, Schl{\"u}ter, and
  Ney}{Sundermeyer et~al\mbox{.}}{2012}]%
        {lstm_lm}
\bibfield{author}{\bibinfo{person}{Martin Sundermeyer}, \bibinfo{person}{Ralf
  Schl{\"u}ter}, {and} \bibinfo{person}{Hermann Ney}.}
  \bibinfo{year}{2012}\natexlab{}.
\newblock \showarticletitle{LSTM neural networks for language modeling}. In
  \bibinfo{booktitle}{\emph{Thirteenth Annual Conference of the International
  Speech Communication Association}}.
\newblock


\bibitem[\protect\citeauthoryear{Sutskever, Vinyals, and Le}{Sutskever
  et~al\mbox{.}}{2014}]%
        {seq2seq}
\bibfield{author}{\bibinfo{person}{Ilya Sutskever}, \bibinfo{person}{Oriol
  Vinyals}, {and} \bibinfo{person}{Quoc~V Le}.}
  \bibinfo{year}{2014}\natexlab{}.
\newblock \showarticletitle{Sequence to Sequence Learning with Neural
  Networks}.
\newblock In \bibinfo{booktitle}{\emph{Advances in Neural Information
  Processing Systems 27}}, \bibfield{editor}{\bibinfo{person}{Z.~Ghahramani},
  \bibinfo{person}{M.~Welling}, \bibinfo{person}{C.~Cortes},
  \bibinfo{person}{N.~D. Lawrence}, {and} \bibinfo{person}{K.~Q. Weinberger}}
  (Eds.). \bibinfo{publisher}{Curran Associates, Inc.},
  \bibinfo{pages}{3104--3112}.
\newblock
\urldef\tempurl%
\url{http://papers.nips.cc/paper/5346-sequence-to-sequence-learning-with-neural-networks.pdf}
\showURL{%
\tempurl}


\bibitem[\protect\citeauthoryear{Tensorflow}{Tensorflow}{2018}]%
        {xla}
\bibfield{author}{\bibinfo{person}{Tensorflow}.}
  \bibinfo{year}{2018}\natexlab{}.
\newblock \bibinfo{booktitle}{\emph{XLA Overview}}.
\newblock
\urldef\tempurl%
\url{https://www.tensorflow.org/performance/xla/}
\showURL{%
\tempurl}


\bibitem[\protect\citeauthoryear{Tran, Bourdev, Fergus, Torresani, and
  Paluri}{Tran et~al\mbox{.}}{2015}]%
        {3d_conv}
\bibfield{author}{\bibinfo{person}{Du Tran}, \bibinfo{person}{Lubomir Bourdev},
  \bibinfo{person}{Rob Fergus}, \bibinfo{person}{Lorenzo Torresani}, {and}
  \bibinfo{person}{Manohar Paluri}.} \bibinfo{year}{2015}\natexlab{}.
\newblock \showarticletitle{Learning Spatiotemporal Features With 3D
  Convolutional Networks}. In \bibinfo{booktitle}{\emph{The IEEE International
  Conference on Computer Vision (ICCV)}}.
\newblock


\bibitem[\protect\citeauthoryear{Truong, Barik, Totoni, Liu, Markley, Fox, and
  Shpeisman}{Truong et~al\mbox{.}}{2016}]%
        {latte}
\bibfield{author}{\bibinfo{person}{Leonard Truong}, \bibinfo{person}{Rajkishore
  Barik}, \bibinfo{person}{Ehsan Totoni}, \bibinfo{person}{Hai Liu},
  \bibinfo{person}{Chick Markley}, \bibinfo{person}{Armando Fox}, {and}
  \bibinfo{person}{Tatiana Shpeisman}.} \bibinfo{year}{2016}\natexlab{}.
\newblock \showarticletitle{Latte: A Language, Compiler, and Runtime for
  Elegant and Efficient Deep Neural Networks}. In
  \bibinfo{booktitle}{\emph{Proceedings of the 37th ACM SIGPLAN Conference on
  Programming Language Design and Implementation}} \emph{(\bibinfo{series}{PLDI
  '16})}. \bibinfo{publisher}{ACM}, \bibinfo{address}{New York, NY, USA},
  \bibinfo{pages}{209--223}.
\newblock
\showISBNx{978-1-4503-4261-2}
\urldef\tempurl%
\url{https://doi.org/10.1145/2908080.2908105}
\showDOI{\tempurl}


\bibitem[\protect\citeauthoryear{Undersander}{Undersander}{2018}]%
        {openai-grad_ckpt-issue}
\bibfield{author}{\bibinfo{person}{Eric Undersander}.}
  \bibinfo{year}{2018}\natexlab{}.
\newblock \bibinfo{booktitle}{\emph{Use with static (unrolled) RNN?}}
\newblock
\urldef\tempurl%
\url{https://github.com/cybertronai/gradient-checkpointing/issues/13}
\showURL{%
\tempurl}


\bibitem[\protect\citeauthoryear{Vasilache, Zinenko, Theodoridis, Goyal,
  DeVito, Moses, Verdoolaege, Adams, and Cohen}{Vasilache
  et~al\mbox{.}}{2018}]%
        {tc}
\bibfield{author}{\bibinfo{person}{Nicolas Vasilache},
  \bibinfo{person}{Oleksandr Zinenko}, \bibinfo{person}{Theodoros Theodoridis},
  \bibinfo{person}{Priya Goyal}, \bibinfo{person}{Zachary DeVito},
  \bibinfo{person}{William~S. Moses}, \bibinfo{person}{Sven Verdoolaege},
  \bibinfo{person}{Andrew Adams}, {and} \bibinfo{person}{Albert Cohen}.}
  \bibinfo{year}{2018}\natexlab{}.
\newblock \showarticletitle{Tensor Comprehensions: Framework-Agnostic
  High-Performance Machine Learning Abstractions}.
\newblock \bibinfo{journal}{\emph{CoRR}}  \bibinfo{volume}{abs/1802.04730}
  (\bibinfo{year}{2018}).
\newblock
\showeprint[arxiv]{1802.04730}
\urldef\tempurl%
\url{http://arxiv.org/abs/1802.04730}
\showURL{%
\tempurl}


\bibitem[\protect\citeauthoryear{Vaswani, Shazeer, Parmar, Uszkoreit, Jones,
  Gomez, Kaiser, and Polosukhin}{Vaswani et~al\mbox{.}}{2017}]%
        {transformer}
\bibfield{author}{\bibinfo{person}{Ashish Vaswani}, \bibinfo{person}{Noam
  Shazeer}, \bibinfo{person}{Niki Parmar}, \bibinfo{person}{Jakob Uszkoreit},
  \bibinfo{person}{Llion Jones}, \bibinfo{person}{Aidan~N. Gomez},
  \bibinfo{person}{{\L}ukasz Kaiser}, {and} \bibinfo{person}{Illia
  Polosukhin}.} \bibinfo{year}{2017}\natexlab{}.
\newblock \showarticletitle{Attention is All you Need}.
\newblock In \bibinfo{booktitle}{\emph{Advances in Neural Information
  Processing Systems 30}}, \bibfield{editor}{\bibinfo{person}{I.~Guyon},
  \bibinfo{person}{U.~V. Luxburg}, \bibinfo{person}{S.~Bengio},
  \bibinfo{person}{H.~Wallach}, \bibinfo{person}{R.~Fergus},
  \bibinfo{person}{S.~Vishwanathan}, {and} \bibinfo{person}{R.~Garnett}}
  (Eds.). \bibinfo{publisher}{Curran Associates, Inc.},
  \bibinfo{pages}{5998--6008}.
\newblock
\urldef\tempurl%
\url{http://papers.nips.cc/paper/7181-attention-is-all-you-need.pdf}
\showURL{%
\tempurl}


\bibitem[\protect\citeauthoryear{Venkataramani, Ranjan, Banerjee, Das, Avancha,
  Jagannathan, Durg, Nagaraj, Kaul, Dubey, and Raghunathan}{Venkataramani
  et~al\mbox{.}}{2017}]%
        {scaledeep}
\bibfield{author}{\bibinfo{person}{Swagath Venkataramani},
  \bibinfo{person}{Ashish Ranjan}, \bibinfo{person}{Subarno Banerjee},
  \bibinfo{person}{Dipankar Das}, \bibinfo{person}{Sasikanth Avancha},
  \bibinfo{person}{Ashok Jagannathan}, \bibinfo{person}{Ajaya Durg},
  \bibinfo{person}{Dheemanth Nagaraj}, \bibinfo{person}{Bharat Kaul},
  \bibinfo{person}{Pradeep Dubey}, {and} \bibinfo{person}{Anand Raghunathan}.}
  \bibinfo{year}{2017}\natexlab{}.
\newblock \showarticletitle{ScaleDeep: A Scalable Compute Architecture for
  Learning and Evaluating Deep Networks}. In
  \bibinfo{booktitle}{\emph{Proceedings of the 44th Annual International
  Symposium on Computer Architecture}} \emph{(\bibinfo{series}{ISCA '17})}.
  \bibinfo{publisher}{ACM}, \bibinfo{address}{New York, NY, USA},
  \bibinfo{pages}{13--26}.
\newblock
\showISBNx{978-1-4503-4892-8}
\urldef\tempurl%
\url{https://doi.org/10.1145/3079856.3080244}
\showDOI{\tempurl}


\bibitem[\protect\citeauthoryear{Wikipedia}{Wikipedia}{2011}]%
        {pcie3_0}
\bibfield{author}{\bibinfo{person}{Wikipedia}.}
  \bibinfo{year}{2011}\natexlab{}.
\newblock \bibinfo{booktitle}{\emph{PCI Express 3.0}}.
\newblock
\urldef\tempurl%
\url{https://en.wikipedia.org/wiki/PCI_Express#PCI_Express_3.0}
\showURL{%
\tempurl}


\bibitem[\protect\citeauthoryear{WMT16}{WMT16}{2016}]%
        {wmt16}
\bibfield{author}{\bibinfo{person}{WMT16}.} \bibinfo{year}{2016}\natexlab{}.
\newblock \bibinfo{booktitle}{\emph{ACL 2016 First Conference on Machine
  Translation}}.
\newblock
\urldef\tempurl%
\url{https://www.statmt.org/wmt16/translation-task.html}
\showURL{%
\tempurl}


\bibitem[\protect\citeauthoryear{Wolfe}{Wolfe}{1990}]%
        {kernel-fusion}
\bibfield{author}{\bibinfo{person}{Michael~Joseph Wolfe}.}
  \bibinfo{year}{1990}\natexlab{}.
\newblock \bibinfo{booktitle}{\emph{Optimizing Supercompilers for
  Supercomputers}}.
\newblock \bibinfo{publisher}{MIT Press}, \bibinfo{address}{Cambridge, MA,
  USA}.
\newblock
\showISBNx{0262730820}


\bibitem[\protect\citeauthoryear{Wu, Schuster, Chen, Le, Norouzi, Macherey,
  Krikun, Cao, Gao, Macherey, Klingner, Shah, Johnson, Liu, Łukasz Kaiser,
  Gouws, Kato, Kudo, Kazawa, Stevens, Kurian, Patil, Wang, Young, Smith, Riesa,
  Rudnick, Vinyals, Corrado, Hughes, and Dean}{Wu et~al\mbox{.}}{2016}]%
        {gnmt}
\bibfield{author}{\bibinfo{person}{Yonghui Wu}, \bibinfo{person}{Mike
  Schuster}, \bibinfo{person}{Zhifeng Chen}, \bibinfo{person}{Quoc~V. Le},
  \bibinfo{person}{Mohammad Norouzi}, \bibinfo{person}{Wolfgang Macherey},
  \bibinfo{person}{Maxim Krikun}, \bibinfo{person}{Yuan Cao},
  \bibinfo{person}{Qin Gao}, \bibinfo{person}{Klaus Macherey},
  \bibinfo{person}{Jeff Klingner}, \bibinfo{person}{Apurva Shah},
  \bibinfo{person}{Melvin Johnson}, \bibinfo{person}{Xiaobing Liu},
  \bibinfo{person}{Łukasz Kaiser}, \bibinfo{person}{Stephan Gouws},
  \bibinfo{person}{Yoshikiyo Kato}, \bibinfo{person}{Taku Kudo},
  \bibinfo{person}{Hideto Kazawa}, \bibinfo{person}{Keith Stevens},
  \bibinfo{person}{George Kurian}, \bibinfo{person}{Nishant Patil},
  \bibinfo{person}{Wei Wang}, \bibinfo{person}{Cliff Young},
  \bibinfo{person}{Jason Smith}, \bibinfo{person}{Jason Riesa},
  \bibinfo{person}{Alex Rudnick}, \bibinfo{person}{Oriol Vinyals},
  \bibinfo{person}{Greg Corrado}, \bibinfo{person}{Macduff Hughes}, {and}
  \bibinfo{person}{Jeffrey Dean}.} \bibinfo{year}{2016}\natexlab{}.
\newblock \showarticletitle{Google's Neural Machine Translation System:
  Bridging the Gap between Human and Machine Translation}.
\newblock \bibinfo{journal}{\emph{CoRR}}  \bibinfo{volume}{abs/1609.08144}
  (\bibinfo{year}{2016}).
\newblock
\urldef\tempurl%
\url{http://arxiv.org/abs/1609.08144}
\showURL{%
\tempurl}


\bibitem[\protect\citeauthoryear{Zaremba, Sutskever, and Vinyals}{Zaremba
  et~al\mbox{.}}{2014}]%
        {rnn_regularize}
\bibfield{author}{\bibinfo{person}{Wojciech Zaremba}, \bibinfo{person}{Ilya
  Sutskever}, {and} \bibinfo{person}{Oriol Vinyals}.}
  \bibinfo{year}{2014}\natexlab{}.
\newblock \showarticletitle{Recurrent neural network regularization}.
\newblock \bibinfo{journal}{\emph{arXiv preprint arXiv:1409.2329}}
  (\bibinfo{year}{2014}).
\newblock


\bibitem[\protect\citeauthoryear{Zheng}{Zheng}{2018}]%
        {mxnet-memory_behavior}
\bibfield{author}{\bibinfo{person}{Jack Zheng}.}
  \bibinfo{year}{2018}\natexlab{}.
\newblock \bibinfo{booktitle}{\emph{Understanding Memory Allocation in MXNet}}.
\newblock
\urldef\tempurl%
\url{https://github.com/apache/incubator-mxnet/issues/11402}
\showURL{%
\tempurl}


\bibitem[\protect\citeauthoryear{Zhu, Akrout, Zheng, Pelegris, Phanishayee,
  Schroeder, and Pekhimenko}{Zhu et~al\mbox{.}}{2018}]%
        {tbd}
\bibfield{author}{\bibinfo{person}{Hongyu Zhu}, \bibinfo{person}{Mohamed
  Akrout}, \bibinfo{person}{Bojian Zheng}, \bibinfo{person}{Andrew Pelegris},
  \bibinfo{person}{Amar Phanishayee}, \bibinfo{person}{Bianca Schroeder}, {and}
  \bibinfo{person}{Gennady Pekhimenko}.} \bibinfo{year}{2018}\natexlab{}.
\newblock \showarticletitle{Benchmarking and Analyzing Deep Neural Network
  Training}. In \bibinfo{booktitle}{\emph{IEEE International Symposium on
  Workload Characterization (IISWC) 2018}}.
\newblock


\end{thebibliography}
